\RequirePackage{fix-cm}
\documentclass[twocolumn]{svjour3}          
\smartqed  
\usepackage{graphicx}

\usepackage{times}
\usepackage{epsfig}
\usepackage{graphicx}
\usepackage{amsmath}
\usepackage{amssymb}
\usepackage{bm}
\usepackage{color}
\usepackage{subfigure}
\usepackage{url}
\usepackage{natbib}

\usepackage{array}

\newcolumntype{M}[1]{>{\centering\arraybackslash}m{#1}}
\newcolumntype{N}{@{}m{0pt}@{}}

\newcommand{\argmax}[1]{\underset{#1}{\operatorname{arg}\,\operatorname{max}}\;}

\begin{document}

\title{Discovering Attribute Shades of Meaning with the Crowd}




\author{Adriana Kovashka \and
        Kristen Grauman
}


\institute{Adriana Kovashka and Kristen Grauman\at
              The University of Texas at Austin\\
              2317 Speedway, Stop D9500, Austin, TX 78712\\
              \email{\{adriana,grauman\}@cs.utexas.edu}
}

\date{Received: date / Accepted: date}

\maketitle


\begin{abstract}
To learn semantic attributes, existing methods typically train one discriminative model for each word in a vocabulary of nameable properties.  However, this ``one model per word'' assumption is problematic: while a word might have a precise linguistic definition, it need not have a precise visual definition.  We propose to discover \emph{shades} of attribute meaning.  Given an attribute name, we use crowdsourced image labels to discover the latent factors underlying how different annotators perceive the named concept.  We show that structure in those latent factors helps reveal shades, that is, interpretations for the attribute shared by some group of annotators.  Using these shades, we train classifiers to capture the primary (often subtle) variants of the attribute.  The resulting models are both semantic and visually precise.  By catering to users' interpretations, they improve attribute prediction accuracy on novel images.  Shades also enable more successful attribute-based image search, by providing robust personalized models for retrieving multi-attribute query results.  They are widely applicable to tasks that involve describing visual content, such as zero-shot category learning and organization of photo collections.
\keywords{Attribute learning and perception \and Vision and language \and Attribute discovery}
\end{abstract}


\vspace{-0.2in}

\section{Introduction}
\label{sec:intro}

\begin{figure*}[t]
\centering
\includegraphics[width=1\linewidth]{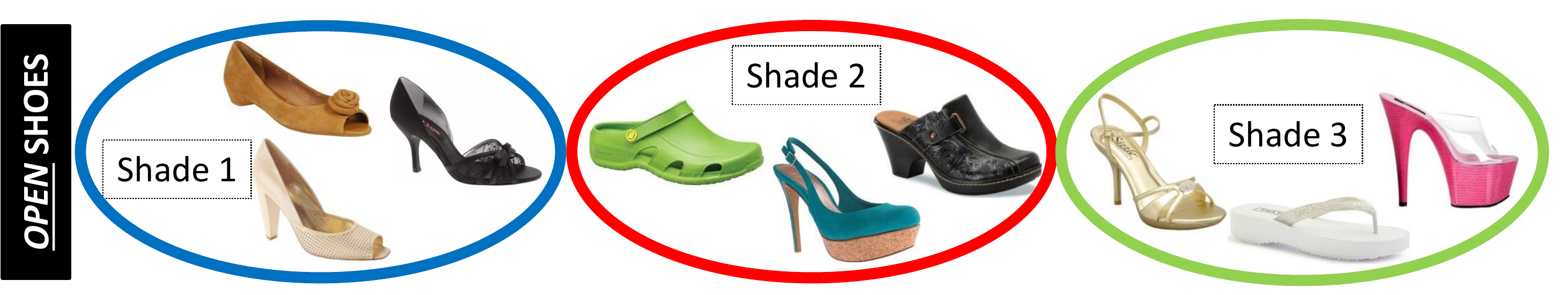}
\vspace{-0.1in}
\caption{Our method uses the crowd to discover factors responsible for an attribute's presence, then learns predictive models based on those visual cues.  For example, for the attribute \emph{open}, our method will discover multiple shades of meaning, e.g., peep-toed (\emph{open} at toe) vs. slip-on (\emph{open} at heel) vs. sandal-like (\emph{open} at toe \emph{and} heel), which are three visual definitions of openness.  Since these shades are not coherent in terms of their global image descriptors, they would be difficult to discover using traditional image clustering. Discovering attribute shades requires both visual cues and semantics.}
\label{fig:concept}
\end{figure*}


Attributes are semantic properties of objects and scenes.  They can
correspond to textures, materials, functional affordances, parts, moods, or
other human-understandable aspects \citep{Ferrari07,Lampert09,Farhadi09,Parikh11b,Kumar11}. 
For instance, a scene can be ``manmade'', or one shoe can be
``more formal'' than another. By injecting language into visual analysis, attributes broaden
the visual recognition problem---from labeling images, to \emph{describing}
them.  This linguistic interpretability opens up several interesting
applications.  For example, a user can search for an image by describing
it \citep{Vaquero09,Kumar11,Siddiquie11,Kovashka12,Scheirer12};
train an object model by describing the category \citep{Lampert09,Parikh11b,Kovashka11,Parkash12}; 
or help the system perform fine-grained recognition by naming the object's
properties \citep{Branson10}.


Typically one defines a vocabulary of attribute words relevant to the domain
at hand---e.g., a vocabulary of facial characteristics for people
search \citep{Kumar11}, textures and parts for
animals \citep{Lampert09,Everingham09,Branson10}, or clothing
properties for shopping \citep{Berg10,Kovashka12}.  Then one gathers labeled
images depicting each attribute in the vocabulary, and trains a model to
recognize each word.


The problem with this standard approach, however, is that there is often a gap between language and visual perception.  In particular, \emph{the words in an attribute vocabulary need not be visually precise.}  An attribute word may connote multiple ``shades" of meaning---whether due to polysemy, variable context-specific meanings, or differences in humans' perception.  For instance, the attribute \emph{open} can describe a door that's ajar, a fresh countryside scene, a peep-toe high heel, or a backless clog.\footnote{Note multiple shades of an attribute may exist even within a specific object category (like shoes, in this example).}  Each shade is distinct and may require dramatically different visual cues to correctly capture.  Thus, the standard approach of learning a single classifier for the attribute as a whole may break down.

Humans often form ``schools of thought'' based on how they interpret and use particular visual attributes. This problem is studied in work on linguistic relativity \citep{Everett13}, which examines how language affects perception and how cultural differences influence how people describe objects, shape properties of animals, colors, etc. 
Colors are the \linebreak quintessential example: e.g., Russian has two words for what would be shades of ``blue'' in English, while other languages do not strongly distinguish ``blue'' and ``green''. 
In other words, if asked whether an object in some image is ``blue'' or not, people of different countries might be grouped around different answers, namely the shades of the attribute.
According to linguistic relativity, speakers of different languages might also exhibit different behavior in tasks involving localization, positioning and classification of objects \citep{Levinson96,Lucy92}.

In addition to language-based factors, attribute use might also differ due to cultural factors. For example, a person who lives in the countryside might have a higher threshold for scene ``naturalness'' or lower threshold for scene ``clutter''. Further, judgments of how ``conservative'' or ``comfortable'' a clothing item is might vary between different countries or even regions within the same country.
For many attributes, such ambiguities in language \emph{use} cannot be resolved by adjusting the attribute definitions, since people \emph{use the same definition differently}.


Unfortunately, neither bottom-up attribute ``discovery" nor relative attributes solve the problem.
Unsupervised discovery methods detect clusters or splits in the low-level
image descriptor space \citep{Parikh11a,Rastegari12,Yu13}. 
While they might discover
finer-grained shades of \emph{some} property, they need not be human-nameable
(semantic).  Furthermore, discovery methods are intrinsically biased by the
choice of features.  For example, the set of salient splits in color
histogram space will be quite different than those discovered in a dense SIFT
feature space.  Similarly, unsupervised methods that cluster global
image descriptors have no way to intelligently focus on only localized regions of
the image, yet an attribute may occupy an arbitrarily small part of an image.


Relative attributes \citep{Parikh11b} do not address the existence of shades, either.  They represent whether an image has a property ``more" or ``less".  
The point in relative attributes is that people may agree best on \emph{comparisons} or \emph{strengths}, not binary labels.
However, just like categorical attributes, relative attributes assume that there is some single, common interpretation of the property shared consistently by all human viewers---namely, that a single ordering of images from least to most [attribute] is possible.  
Thus, shades are relevant whether the attributes are modeled with classifiers (binary) or ranking functions (relative).


Our goal is to automatically discover the shades of an attribute.
\textbf{An attribute shade is a visual interpretation of an attribute name that one or more people apply when judging whether that attribute is present in an image.}
Similarly, if learning relative attributes, a shade is an interpretation when judging whether that attribute is present more in image A or image B. 
See Figure~\ref{fig:concept}.

Given a semantic attribute name, we want to discover its multiple visual
interpretations and train a discriminative model for each one.  Rather than
attempt to manually enumerate the possible shades, we propose to learn them
indirectly from the crowd.  First we ask many annotators to label various
images, reporting whether the attribute is present or not.  Using their
responses, we estimate latent factors that represent the annotators in terms
of the kinds of visual cues that they associate with the attribute. Then,
clustering in the low-dimensional latent space, we identify the schools of thought
(about how to interpret this attribute) underlying the discrete set of labels the annotators provided.
(We use the terms ``school'' and ``shade'' interchangeably.)
Finally, we use the positive exemplars in each school to train a predictive
model, which can then detect when the particular attribute shade is present
in novel images.


The resulting models are both semantic and visually precise.  By discovering
the shades from the crowd's latent factors, we isolate the features
corresponding to the perceived shades.  This makes our method less
susceptible to the more ``obvious" splits in the feature space that an image
clustering approach---including today's sophisticated discovery
methods such as \citep{Parikh11a,Rastegari12,Yu13}---may find, which need not
directly support the semantic attribute of interest.  

Note that work in automatically finding the multiple senses of a polysemous word \citep{Barnard06b,Loeff06,Saenko08,Berg06} is orthogonal to our goal, as it focuses on nouns (object categories), not descriptive properties.  Further, the visual differences of polysemous nouns are usually stark (e.g., a river \emph{bank} or financial \emph{bank}).  In contrast, attribute shades are often subtle differences in interpretation.
We study the problem of automatically discovering shades of adjectives, and determining which shade of an adjective a user employs when judging whether a visual property is present or not in a particular image. 


On two datasets, we find that not only are the discovered shades visually meaningful, they are also well-aligned with annotators' textual explanations of their labels.  Most importantly, we show their practical utility to reliably estimate perceived attributes in novel images, which is crucial for any application relying on the descriptive nature of attributes (e.g., image search or zero-shot learning).

\section{Related Work}
\label{sec:related}

\vspace{-0.1in}
\paragraph{Learning attributes}

Attributes are nameable visual properties that can aid both classification \citep{Lampert09,Farhadi09,Branson10,WangY10,Parikh11b,Patterson12} and image search \citep{Kumar11,Vaquero09,Kovashka12,Siddiquie11,Scheirer12}.  Whether categorical or relative, prior work assumes that each attribute word corresponds to one coherent visual property, and so trains one classifier \citep{Ferrari07,Kumar11,Lampert09,Farhadi09,Vaquero09,Branson10,WangY10,Patterson12} or one ranking function \citep{Parikh11b,Kovashka12} per attribute.

Since annotators may disagree about the attribute label for an
image \citep{Farhadi09,Endres10,Patterson12,Curran12}, the norm is to take
the majority vote label (and discard the image if votes are too split). Thus,
prior work treats differences in attribute perception as noise.  To our
knowledge, the only exception is our transfer learning approach
\citep{Kovashka13b}, which trains \emph{user-specific} models for
personalized image search.  
In that work, we adapt a generic model for an attribute using training data from each individual user,
and the method produces one attribute classifier for each user.
In contrast, in this work we discover schools of thought among
the crowd, and our method produces a set of attribute shades capturing
commonly perceived variations.  These schools of thought are a valuable midpoint on the spectrum from purely consensus models to purely user-specific models, resulting in better accuracy for perceived attributes (cf. Sec.~\ref{sec:models}).  Shades also have broader utility than the adapted user-specific models \citep{Kovashka13b}, since they let us explicitly organize perceived properties.

\vspace{-0.1in}
\paragraph{Distinction with relative attributes}

We stress that relative attributes \citep{Parikh11b}, while avoiding the need for forced categorical judgments, still assume a single underlying visual property exists.  They do not represent multiple interpretations.  
For example, relative attributes construct a \emph{universal} model for ``less brown" vs. ``more brown".  
They do not address the issue that one person may say ``image X is \emph{browner} than Y'', while another may say the opposite.
Shades, on the other hand, are concerned with discovering \emph{multiple} models for varying perceptions of brown, e.g., chocolate brown vs. goldish brown.  The two goals are orthogonal.  
In fact, while we study categorical attributes, the proposed approach could easily be applied to discover shades of relative attributes; the label matrix in Sec.~\ref{sec:bpmf} would simply record whether the person finds a first image to exhibit the attribute more or less than a second image.

\vspace{-0.1in}
\paragraph{Defining attribute vocabularies}

Most work defines the attribute vocabulary manually, or by eliciting
discriminative properties from
annotators \citep{Patterson12,Maji12}.  However, in some
cases it is possible to generate it
(semi-)automatically, as in \citep{Everingham09,Branson10,Berg10,Parikh11a,Rohrbach12}.
For animal species, field guides are a natural source of attribute
names \citep{Everingham09,Branson10}.  Given their focus on concrete
parts, such domains are less prone to shades.  When suitable text sources are
available---such as captioned images on web pages \citep{Berg10} or activity
scripts \citep{Rohrbach12}---one can mine for candidate attribute words.  
Since not all words will be visually detectable, some work aims to prune the vocabulary automatically \citep{Berg10, Barnard06a}. 
Rather than mined text, our shades use sparse crowd labels to capture latent interpretations of an attribute,
which may not be concisely describable with a keyword.

\vspace{-0.1in}
\paragraph{Discovering non-semantic attributes}

While the term ``attribute" typically connotes a \emph{semantic} property,
some researchers also use the term to refer to discovered \emph{non-semantic}
features \citep{Mahajan11,Rastegari12,Sharmanska12,Yu13}.  The idea is to
identify ``splits" or clusters in the low-level image descriptor space, often
subject to constraints that deter redundancy and promote discriminativeness
for object recognition.  However, being bottom-up, there is no guarantee the
splits will correspond to a nameable property.  Hence, unlike our shades,
they are non-semantic and inapplicable to descriptive attribute tasks, like image search or zero-shot learning.  One
can attempt to assign names to discovered ``attributes" after the
fact, as in \citep{Parikh11a,Duan12,Yu13}, but the patterns that are even
discoverable remain biased by the chosen low-level image feature space, as
discussed above.
Semantics and human interpretability are essential if human users are to use attributes to communicate with a vision system.

\vspace{-0.1in}
\paragraph{Polysemy and domain adaptation}

A polysemous word has multiple ``senses" or meanings.  Some work bridging
text and visual analysis aims to cluster Web images according to distinct
senses \citep{Barnard06b,Loeff06,Saenko08,Berg06}.  
Other approaches find within-category modes in order to perform better domain adaptation for object recognition \citep{Hoffman12,Gong13,Xiong14}.  
These works are orthogonal to our goal, as they focus on nouns and object categories, not descriptive properties.  
Typically the visual differences between senses of a polysemous word (or surrounding text context) are much larger than between attribute shades of meaning.  
Distinctions between attributes, on the other hand, are more subtle, and they are tied to semantics more so than to visual differences.  
Furthermore, unlike a truly polysemous word, for which one can enumerate the multiple dictionary definitions, attribute shades are often more difficult to definitively
express in language.  We show how to automatically infer them from trends in crowd labels.

\vspace{-0.1in}
\paragraph{Aggregating crowd labels}

Crowd input has been aggregated in novel ways for image clustering \citep{Gomes11}, image similarity \citep{Tamuz11}, and object labeling \citep{Welinder10}.  Welinder et al. model annotators' competence and bias to discover their schools of thought, and subsequently undo their biases to produce more reliable ground truth.  While that work aims to recover a single true label for each image, our goal is to discover the crowd's multiple interpretations of a label.

Our method makes use of an existing matrix factorization algorithm \citep{Sal08}.
Matrix factorization is often used for matrix completion, to solve
collaborative filtering problems (e.g., the Netflix challenge) by exploiting
commonalities among users \citep{Sal08, Xiong10}.  Rather than impute missing
labels, we propose to use the latent factors themselves to represent the
interplay between language, human perception, and image examples.  
Furthermore, we show how to use the recovered schools of thought to build content-based attribute models.

\section{Approach}
\label{sec:approach}

In order to discover shades of attributes, we first recover the latent factors that motivate a user's annotations of an image with a given attribute's presence or absence. We then represent each user in this latent space, and discover groupings among users. Each group or school is mapped to the images which are most frequently believed to contain the attribute, according to the corresponding shade of the attribute. Using these images, we learn models that predict whether the attribute is present or not in a novel image, for some school/shade.

We first explain the crowdsourced label collection in Section \ref{sec:col}.  Then we describe how we recover the latent factors responsible for those labels (Section \ref{sec:bpmf}) and use them to discover attribute shades (Section \ref{sec:schools}).  Finally, we exploit the discovered shades to improve attribute prediction by accounting for the users' varying interpretations (Section \ref{sec:models}).

\subsection{Collecting Crowd Labels per Attribute}
\label{sec:col}

We use two datasets: \textbf{Shoes} \citep{Berg10, Kovashka12} and \textbf{SUN Attributes} \citep{Patterson12}.  
While attribute labels are available for both, our method needs to record which annotator labeled which image.  Thus, we run our own crowdsourced label collection.

To focus our study on plausibly ``shaded" words, we select 12 attributes that can be defined concisely in language, yet may vary in their visual instantiations.
This helps ensure that variance in the annotators' labels stems from the attribute's visual sub-meanings, as opposed to external factors like the annotator's personal taste.  The 12 attributes are: ``pointy'', ``open'', ``ornate'', ``comfortable'', ``formal'', ``fashionable'', ``brown'' (for Shoes); and ``cluttered'', ``soothing'', ``open area'', ``modern'', ``rustic'' (for SUN).  
We obtain definitions of the attributes from a web dictionary, and show these in Table \ref{tab:defs}.

In general, we choose words whose application in conversation requires some interpretation of the definition. This interpretation can revolve around judging thresholds and establishing what factors cause the definition to hold. For example, for the ``open area'' attribute, one is required to judge what constitutes ``unobstructed passage''; for ``open'', how many (and how big) gaps there are; for ``ornate'', which patterns matter; for ``comfortable'', what aspect of the shoe causes comfort. We also choose words whose presence or absence involves personal knowledge or beliefs; e.g., for ``rustic'', one should determine what country life is like.

Our decision to focus on words likely to have shaded meanings lets us examine the problem at hand most directly.  However, even if some attributes in the pool turn out to be fairly precise visually, our method is capable of returning few shades or just one shade, since we employ automatic model selection. Thus, applying the shades discovery algorithm we propose to a ``less shaded'' word should in principle do no harm. 

We sample  $N=250$ to $1000$ images per attribute.  To get representative images spanning the dataset, we cluster all images using $K$-means, then sample ones near the cluster centers.\footnote{For ``brown'', we sample images with high scores output by a ``brown'' classifier.  This attribute is rare, so sampling cluster centers would produce very few brown images.} 
This yields a total of 2559 images for Shoes and 2086 images for SUN.

\begin{table}[t]
\centering
\begin{tabular}{|M{2cm}|M{5.3cm}|N}
\hline
Attribute & Dictionary definition &\\[12pt]
\hline
Pointy & having a comparatively sharp point, or having numerous pointed parts&\\[12pt]
Open & having interspersed gaps, spaces, or intervals&\\[12pt]
Ornate & made in an intricate shape or decorated with complex patterns&\\[12pt]
Comfortable &  providing physical comfort, ease and relaxation&\\[12pt]
Formal & designed for wear or use at elaborate ceremonial or social events&\\[12pt]
Brown & the color of, for example, chocolate and coffee&\\[12pt]
Fashionable & conforming to the current fashion; stylish; trendy; modern&\\[12pt]
\hline
To clutter & to make disorderly or hard to use by filling or covering with objects&\\[12pt]
To soothe & to bring comfort, composure, or relief&\\[4pt]
Open (area) & affording unobstructed passage or view&\\[4pt]
Modern & characteristic or expressive of recent times or the present; contemporary &\\[12pt]
Rustic & of, relating to, or typical of country life or country people&\\[12pt]
\hline
\end{tabular}
\caption{The 12 attribute definitions shown to annotators.}
\label{tab:defs}
\end{table}

We build a Mechanical Turk interface to gather the labels.  Workers are shown definitions of the attributes (Table \ref{tab:defs}) as part of the task instructions.  These instructions are visible during task completion.  However, workers are shown no example images.  Thus, they all receive the same linguistic definition, but they are not prompted with any particular \emph{visual} definition.  Then, given an image, the worker must provide a binary label, i.e., he or she must state whether the image does or does not possess a specified attribute.  Additionally, for a random set of five images, the worker must explain his label in free-form text, and state which image most has the attribute, and why.  These questions both slow the worker down, helping quality control, and also provide valuable ground truth data for evaluation, as we will explain in Section \ref{sec:coherency}.  

Our latent factor model (defined next) can accommodate imbalanced and sparse labels.  This is good, because in realistic scenarios, labels may not originate from concentrated one-time labeling efforts (like ours), but rather as a side product of another task---such as click data in image search.  In such a case, the images that one user labels will not entirely overlap with those that another user labels.  Furthermore, each user will label few examples.  To mimic this scenario, we gather labels in a sparse fashion.  Each worker labels 50 randomly chosen images, per attribute.  
To help ensure self-consistency in the labels, we exclude workers who fail to consistently answer three repeated questions sprinkled among the 50.  
This yields annotations from 195 workers per attribute on average.

While multiple workers may label the same image, we stress their labels are \emph{not} aggregated to create a majority vote ``ground truth".  The main premise of shades is that attribute names can be visually imprecise and so admit multiple interpretations.  The same attribute word can have different meanings to different people, even if they all know the same linguistic definition of the word.  (Contrast this with object category names, which are relatively precise.)   Thus, rather than discard label discrepancies as noise, we use them to discover shades.

\subsection{Recovering Latent Factors for Attribute Labels}
\label{sec:bpmf}

Now we use the label data to discover latent factors, which are needed to recover the shades of meaning.  Note that we learn factors for each attribute independently, so all variables below are attribute-specific.  From the above data collection, we retain each worker's ID, the indices of images he labeled, and how he labeled them.  Let $M$ denote the number of unique annotators, and let $N$ denote the number of images seen by at least one annotator.  Let $\mathbf{L}$ be the $M \times N$ label matrix, where $L_{ij} \in \{0,1,?\}$ is a binary attribute label for image $j$ by annotator $i$.  A ? denotes an unlabeled example.  The matrix is only partially observed, as on average only 20\% of the possible image-worker pairs are labeled.

We suppose there is a small number $D$ of unobserved factors that influence the annotators' labels.  This reflects that their decisions are driven by some mid-level visual cues.  For example, when deciding whether a shoe looks ``ornate", the latent factors might include presence of buckles, amount of patterned textures, material type, color, and heel height; when deciding whether a scene looks ``modern", they might include color, object composition, and materials.

Assuming a linear factor model, the label matrix $\mathbf{L}$ can be factored as the product of an $M \times D$ annotator latent factor matrix $\mathbf{A}^T$ and a $D \times N$ image latent factor matrix $\mathbf{I}$: 
\begin{equation}
\mathbf{L} = \mathbf{A}^T \mathbf{I}.
\end{equation}

A number of existing methods can be used to factor this partially observed matrix, by finding the best rank-$D$ approximation under some loss function \citep{Sal07,Sal08,Xiong10}.  We use a probabilistic matrix factorization algorithm (PMF) from \citep{Sal07,Sal08}, due to its efficiency for large, sparse matrices.  Briefly, it works as follows.  PMF takes a probabilistic approach to recover the two low-rank matrices.  
Let $A_i$ and $I_j$ denote columns of $\mathbf{A}$ and $\mathbf{I}$, respectively, and $\ell_{ij}=1$ if we received a label on image $j$ by annotator $i$, and $\ell_{ij}=0$ otherwise.
The likelihood distribution for the observed labels is 
\begin{equation}
p(\mathbf{L} | \mathbf{A}, \mathbf{I}, \sigma^2) = \prod_{i=1}^M \prod_{j=1}^N \big[\mathcal{N}(L_{ij} | A_i^T I_j, \sigma^2)\big]^{\ell_{ij}},
\label{eqn:shades}
\end{equation}
where 
$\mathcal{N}(x | \mu, \sigma^2)$ denotes a Gaussian distribution with mean $\mu$ and standard deviation $\sigma^2$.  The priors over the latent factors are spherical Gaussians:
\begin{equation}
p(\mathbf{A} | \sigma_A^2) = \prod_{i=1}^M \mathcal{N}(A_i | 0, \sigma_A^2 \mathbb{I}), \text{and}
\end{equation}
\begin{equation}
p(\mathbf{I} | \sigma_I^2) = \prod_{j=1}^N \mathcal{N}(I_j | 0, \sigma_I^2 \mathbb{I}).
\end{equation}

\begin{figure*}[t]
\centering
\includegraphics[width=1\linewidth]{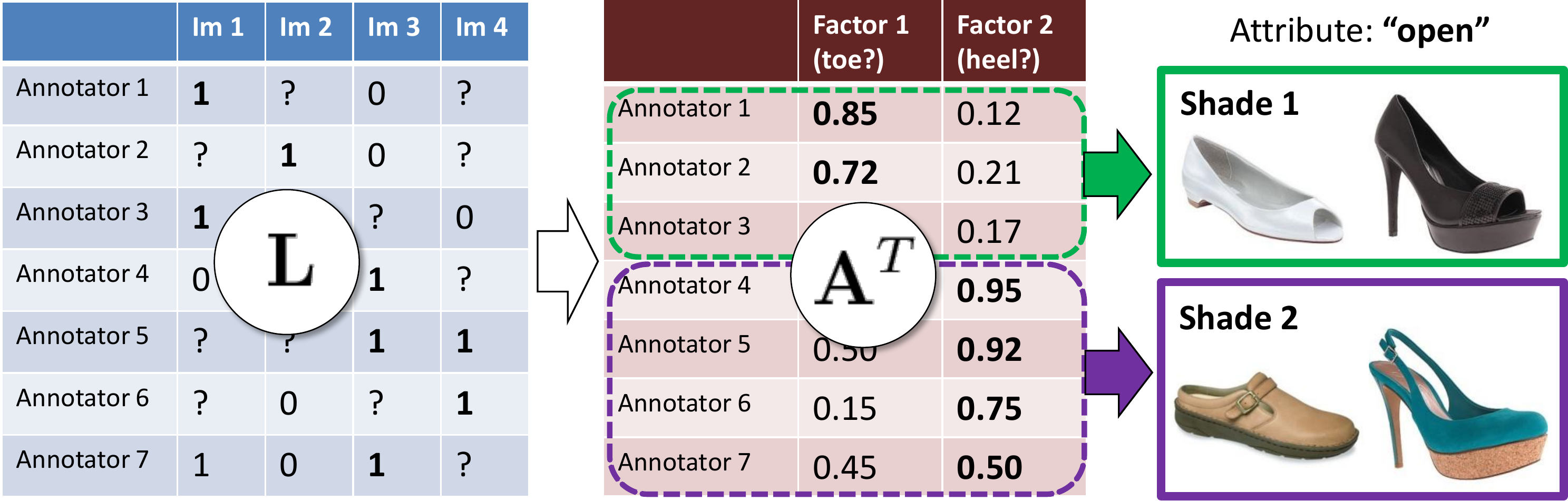}
\caption[Illustration of my attribute shade discovery approach]{Given a partially observed attribute-specific label matrix (left), we recover its latent factors and their influence on each annotator (middle).  We discover shades by clustering in this space (dotted lines in center and images on right).}
\label{fig:approach}
\end{figure*}

We seek the latent features that maximize the log-posterior:
\begin{equation}
\mathbf{A}^\ast, \mathbf{I}^\ast = \argmax{\mathbf{A},\mathbf{I}} \ln p(\mathbf{A},\mathbf{I} | \mathbf{L},\sigma^2, \sigma_A^2, \sigma_I^2).
\end{equation}
Obtaining the MAP factors amounts to minimizing an SSD objective function with quadratic regularization terms using gradient descent \citep{Sal07}:
\begin{equation}
E = \frac{1}{2} \sum_{i=1}^M \sum_{j=1}^N \ell_{ij} (L_{ij} - A_i^T I_j)^2 + \frac{\lambda_A}{2} \sum_{i=1}^M \|A_i\|^2 \nonumber
\end{equation}
\begin{equation}
~~~~~ + \frac{\lambda_I}{2} \sum_{j=1}^N \|I_j\|^2,
\end{equation}
where $\lambda_A = \sigma^2 / \sigma^2_{A}$ and $\lambda_I = \sigma^2 / \sigma^2_{I}$, and we use the Frobenius norm.    

This approach is a probabilistic extension of what would be standard SVD in the case of fully observed labels.  However, performance might depend on careful tuning of parameters such as $\sigma^2$, $\sigma_A^2$, $\sigma_I^2$. 
Upgrading to a full Bayesian treatment \citep{Sal08}, we put priors on the user and image hyperparameters. 
Let the mean and precision matrix of the user and image prior distributions be denoted by $\mu_A$ and $\mu_I$, and $\mathbf{\Lambda}_A$ and $\mathbf{\Lambda}_I$, respectively, and let $\Theta_A = \{\mu_A, \mathbf{\Lambda}_A\}$ and $\Theta_I = \{\mu_I, \mathbf{\Lambda}_I\}$. We place Gaussian-Wishart priors on these hyperparameters $\Theta_A$ and $\Theta_I$:
\begin{equation}
p(\Theta_A | \Theta_0) = \mathcal{N}(\mu_A | \mu_0, (\beta_0 \mathbf{\Lambda}_A)^{-1}) \mathcal{W}(\mathbf{\Lambda}_A | \mathbf{W}_0, \nu_0), \text{and} \nonumber
\end{equation}
\begin{equation}
p(\Theta_I | \Theta_0) = \mathcal{N}(\mu_I | \mu_0, (\beta_0 \mathbf{\Lambda}_I)^{-1}) \mathcal{W}(\mathbf{\Lambda}_I | \mathbf{W}_0, \nu_0),
\end{equation}
where $\Theta_0 = \{\mu_0, \nu_0, \mathbf{W}_0\}$, $\mu_0 = 0$, $\beta_0 = 1$, $\nu_0 = D$, and $\mathbf{W}_0$ is the identity matrix.  

Imputing $L_{ij}$ for some unknown labeling of user $i$ and image $j$ is then predicted via MCMC:
\begin{equation}
p(L^*_{ij}|\mathbf{L}, \Theta_0) \approx \frac{1}{R} \sum_{r=1}^R p(L^*_{ij} | A^{(r)}_i, I^{(r)}_j),
\end{equation} 
where the samples $\{A^{(r)}_i, I^{(r)}_j\}$ are generated in parallel via Gibbs sampling as:
\begin{equation} 
A_i^{(r+1)} \sim p(A_i | \mathbf{L}, \mathbf{I}^{(r)}, \Theta_A^{(r)}), \text{and} \nonumber
\end{equation}
\begin{equation}
I_j^{(r+1)} \sim p(I_j | \mathbf{L}, \mathbf{A}^{(r+1)}, \Theta_I^{(r)}).
\end{equation}
We obtain our estimates of $\mathbf{A}$ and $\mathbf{I}$ by averaging the $R$ samples for each.

This Bayesian treatment reduces overfitting and saves parameter tuning. See \citep{Sal08} for details.

\subsection{Discovering Shades of Meaning}
\label{sec:schools}

In collaborative filtering, the goal of the factorization described above is to impute missing labels (e.g., to predict how a user will rate an unseen movie, $L_{ij} \approx \langle A_i, I_j \rangle$).  While missing labels could similarly be estimated for our data, our goal is different.  We aim to discover attribute shades of interpretation and generate predictive visual models for them.

To this end, we first represent each annotator in terms of his association with each discovered factor.  The ``latent feature vector" for annotator $i$ is $A_i \in \Re^D$, the $i$-th column of $\mathbf{A}$.  It represents how much each of the $D$ factors influences that annotator when he decides if the named attribute is present.  Likewise, the latent feature for image $j$ is $I_j \in \Re^D$, the $j$-th column of $\mathbf{I}$, and represents how much each of the $D$ factors is visible in the image.

Figure~\ref{fig:approach} illustrates with a cartoon example.  As seen on the left, annotators did not label all images for the attribute ``open".  Some tended to label images 1 and 2 as having the attribute, whereas others tended to label 3 and 4 as positive.  After factoring the label matrix, suppose we discover $D=2$ latent factors.  Though nameless, they align with semantic visual cues; suppose here they are ``toe is open" and ``heel is open".  Each annotator's feature $A_i$ encodes how important those two factors were for his label decision.  In this hypothetical example, we see the first three annotators labeled images 1 and 2 as open due to factor 1, whereas the others focused on factor 2 in other images.

We pose shade discovery as a grouping problem in the space of these latent features.\footnote{Though we can cluster either annotators or images to identify shades, we choose annotators in order to facilitate the mapping of users to shades when building predictive models for the shades, as described in Section \ref{sec:models}.}
While various clustering algorithms could be used, we apply $K$-means to the columns of $\mathbf{A}$ to obtain clusters $\{\mathcal{S}_1,\dots,\mathcal{S}_K\}$.\footnote{Preliminary tests with Bayesian non-parametric clustering showed inferior results.  An alternative would be to impute missing labels and group with EM, but clustering in the compact latent space is preferable when labels are very sparse.}  Each cluster is a shade.  Annotators in the same cluster display similar labeling behavior, meaning they interpret similar combinations of mid-level visual cues as salient for the attribute at hand. For example, in Figure \ref{fig:approach}, the two dominant shades reflect which part of the shoe the annotator focused on to judge openness---toe or heel.  (Of course, for real data, there will be $D > 2$ factors, and shades will combine many such factors.)

Recall that shade discovery is done on a per-attribute basis.  Depending on the visual precision of the word, some attributes may have only one shade; others may have many.  To automatically select $K$ based on the structure of the data, we use a variant of the silhouette coefficient \citep{Rousseeuw87}. 
It quantifies the quality of a clustering, by measuring how tightly grouped the latent features in a cluster are, normalized by how far they are from other clusters. More specifically, let $a_i$ be the average Euclidean distance of a cluster member $i$ to its neighbors (members of the same cluster), and let $b_i$ denote the mean distance of $i$ to other clusters, where the distance to each cluster is measured as an average over distances to the cluster members. Then let:
\begin{equation}
s_i = \frac{b_i - a_i}{\max(a_i, b_i)}.
\end{equation}
The silhouette coefficient is computed as the mean of the values $s_i$.

As discussed above, by using automatic model selection, our approach is free to decide that an input word is already visually precise, not requiring many shades.

\subsection{Using Shades to Predict Perceived Attributes}
\label{sec:models}

A key valuable application of shades is to improve attribute prediction accuracy, generalizing what the system discovered to novel images.  

Any method leveraging the descriptive nature of attributes needs to rely on attribute models that match a human user's perception. For example, an image search system that allows attribute-based queries \citep{Kumar11,Siddiquie11,Scheirer12,Rastegari13,Vaquero09,Kovashka12} will frustrate a user if the system's notion of ``formal'' does not match the user's notion. 
Similarly, a zero-shot object recognition system that trains a new object model based on its attribute specification will fail unless it correctly interprets the visual meaning intended by the human teacher.

Prior work uses one of two extremes for attribute prediction---either (1) a \emph{consensus} classifier: a single generic model trained with examples whose labels are obtained through a majority vote over multiple redundant crowd responses (e.g., \citep{Kumar11,Lampert09,Farhadi09,Vaquero09,Patterson12}), or (2) a \emph{user-specific} classifier which is trained by adapting that majority vote model to satisfy an individual user's training labels \citep{Kovashka13b}. In the latter approach, we collect between 12 and 40 labels per attribute from each user, and apply them to learn a user-specific attribute model, which we regularize with the parameters of the generic model.

Shades offer an approach in between these two extremes.  With shades, we can account for the fact that people perceive an attribute differently, yet avoid specializing predictions down to the level of each individual user.  The idea is to tailor an attribute classifier according to the user's ``school of thought", i.e., the shade to which he subscribes.

\begin{figure}[t]
\centering
\includegraphics[width=1\linewidth]{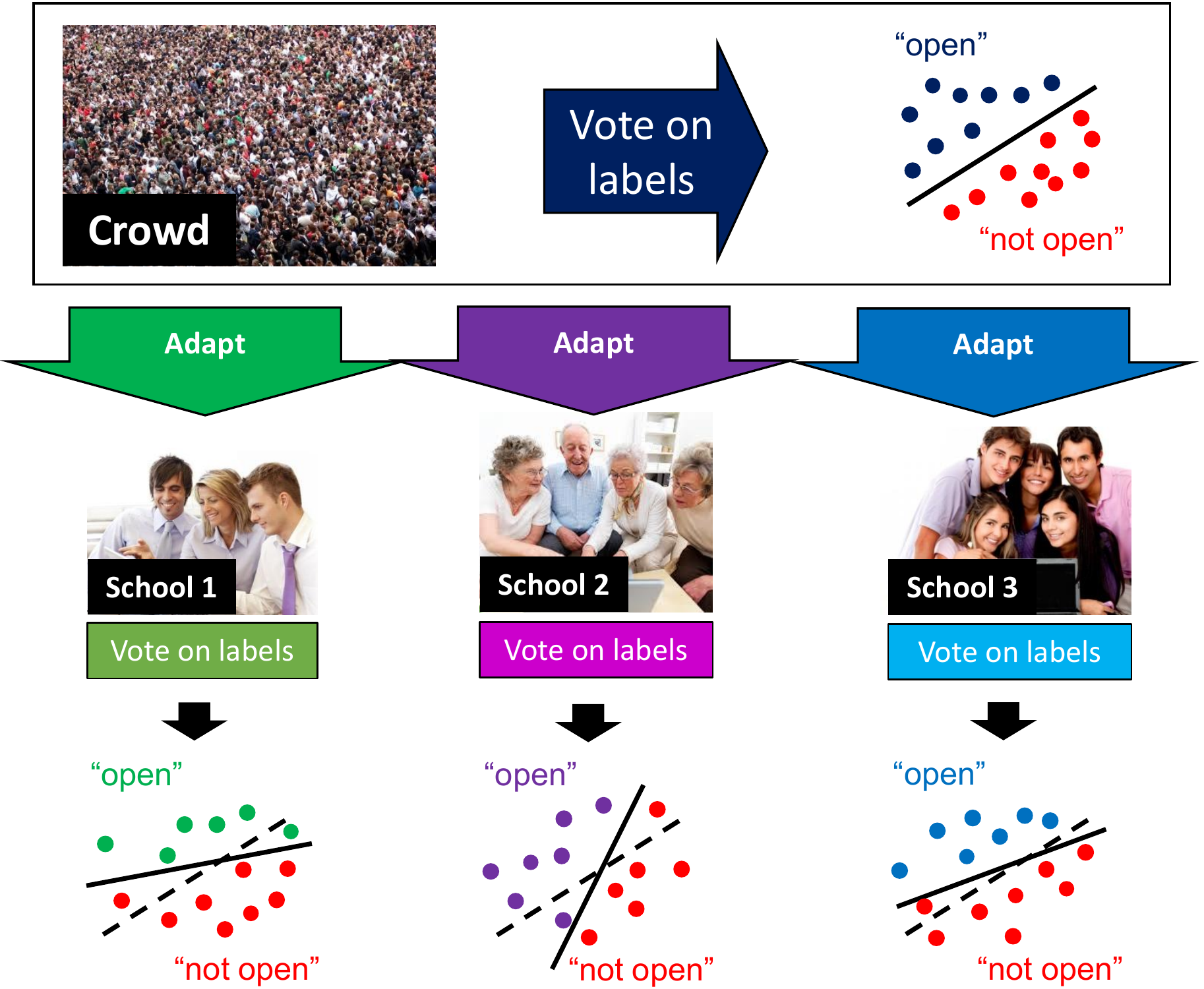}
\caption{We learn predictive models for shades, by adapting a standard consensus model trained from any users in the crowd towards particular schools of users.}
\label{fig:adapt_schools}
\end{figure}

To exploit the existence of schools of thought, we train shade-specific classifiers that adapt the consensus model.  See Figure \ref{fig:adapt_schools}.  Each shade $\mathcal{S}_k$ is represented by the total pool of images that its annotators labeled as positive.  Several annotators in the cluster may have labeled the same image, and their labels need not agree.  Thus, we perform majority vote over just the annotators in $\mathcal{S}_k$ to decide whether an image is positive or negative for the shade.  
This majority vote is a form of quality control, where we assume consistency within the group.  
For both the shade models and the consensus model, we discard labels where fewer than 90\% of users agree.  

We use the resulting image-label pairs to train a discriminative classifier, using the adaptive support vector machine (SVM) objective of \cite{Yang07} to regularize its parameters to be similar to those of the consensus model.  
In other words, we are now personalizing to schools of users, as opposed to individual users.
See Figure \ref{fig:adapt_schools} for an overview of this procedure.
Then we apply the adapted shade model for the cluster to which a user belongs to predict the presence/absence of the attribute in novel images.  Thus, the predictions are automatically tailored to that user's perception of the property.

To recap, shades offer an important midpoint on the spectrum discussed above.  Compared to the standard consensus approach, we account for distinct perceived shades.  Compared to user-adaptive models, the advantages are twofold.  First, each model typically leverages more training data than a single user provides.  This lets us effectively ``borrow" labeled instances from the user's neighbors in the crowd.  Second, we leverage the robustness of the intra-shade majority vote.  This helps reduce noise in an individual user's labeling.  The results in Section \ref{sec:schools-result} reveal the impact of these advantages in practice.

Note, a user must provide at least some attribute labels to benefit from the shade models, since we need to know which shade to apply.  For users who contributed to the label matrix $\mathbf{L}$ this is straightforward.  For users adding labels later, we could either re-factor $\mathbf{L}$, or more efficiently, use a folding-in heuristic~\citep{Dumais90,Hofmann99} (not attempted in our experiments).

\begin{figure*}[t]
\centering
\includegraphics[width=1\linewidth]{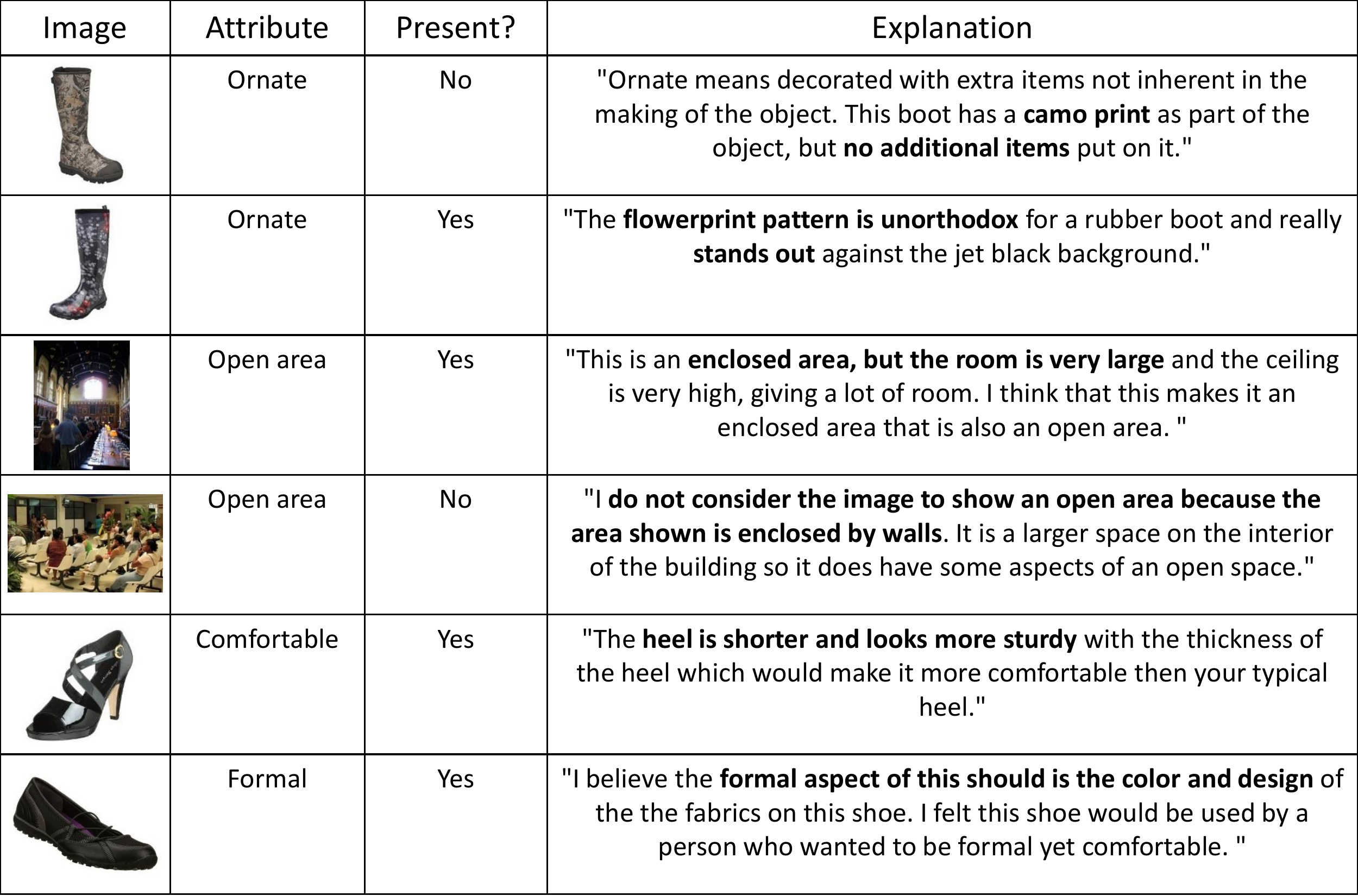}
\caption{Example label explanations that annotators provided. Bold is our emphasis. In the first two rows, notice that the same type of shoe (one with patterns) can be perceived to have a different level of ornamentation, depending on whether the annotator believes patterns constitute ornamentation. Further, a room with large spaces (rows 3 and 4) can be perceived as an open area or not, depending on whether the annotator believes an area enclosed by walls can be considered open. Finally, in the last two rows we see two interesting examples of a high-heeled shoe (which is normally labeled as uncomfortable) considered comfortable due to its sturdy heel, and a sneaker-like shoe seen as formal due to its color and design. Also notice how well-thought out these user responses are, which indicates that the quality of data we collected is high.}
\label{fig:descr}
\end{figure*}

\subsection{Discussion}
\label{sec:discussion}

The key thing to note about the shade classifiers is how their positive labeled exemplars came about.  Images within a shade can be visually diverse from the point of view of typical global image descriptors, since annotators attuned to that shade's latent factors could have focused on arbitrarily small parts of the images, or arbitrary subsets of feature modalities (color, shape, texture).  
For example, one shade for ``open'' might focus on shoe toes, while another focuses on shoe heels.  Similarly, one shade for ``formal'' capturing the notion that dark-colored shoes are formal would rely on color alone, while another capturing the notion that shoes with excessively high heels are not formal would rely on shape alone.  
An approach that attempts to discover shades based on image clustering---or non-semantic attribute discovery such as \citep{Parikh11a,Mahajan11,Duan12,Rastegari12,Sharmanska12,Yu13}---will be hard pressed to group images according to these perceived, possibly subtle, cues.  Our insight is to leverage patterns among the crowd labels to partition the images \emph{semantically}.  Then, even though the training images may be visually diverse, standard discriminative learning methods let us isolate the informative features.  Essentially, we avoid biasing the shades to a particular low-level descriptor space, since their training images are determined independent of the descriptors.

One might wonder: why not just manually enumerate the attribute shades with words?  Our approach has multiple advantages over that strategy, beyond being automatic.  For polysemous \emph{nouns}, the visual definitions are enumerable---one could simply check the dictionary.  
In contrast, it can be difficult to put an attribute's distinct visual instantiations in words, e.g., by automatically generating all possible qualifiers for an attribute.  This would amount to automatically listing all possible contexts in which an object can occur, all possible shapes a human body can take, etc.  
Neither can we rely solely on mining the textual explanations gathered from users to qualify attributes.  
We find that the words annotators typically provide to explain their interpretation of an attribute are concrete instances of the shade, which need not comprehensively define the shade.  For example, in our data collection, when asked to explain why an image is ``ornamented", an annotator might comment on the ``buckle" or ``bow"; yet the latent shade of ``ornamented'' underlying many users' labels is more abstract.  It encompasses combinations of such concrete mid-level cues.  In short, we find that people are good at naming examples, but less good at characterizing an entire shade in words.  
Our method fills that gap, using structure in the labels to identify shades.

Shades require no additional labeling effort compared to the existing user-specific approach \citep{Kovashka13b}.  Yet, by relying on a user's ``neighbors'' in the crowd, we utilize data the user has not labeled but neighbors \emph{have} labeled, thus reducing the manual annotation effort.  In terms of computational complexity, the only added cost compared to the method of \citep{Kovashka13b} is running the Bayesian PMF method, which requires about 21 minutes per attribute (see Section \ref{sec:shades_exp_design}).  Therefore, our shade formation approach offers numerous advantages over alternative approaches, for only a small complexity overhead.

\begin{table*}[t]
\centering
\begin{tabular}{|c||c|c|c|c|c|c|}
\hline
Attribute & {\scriptsize \textsc{Shades}} & {\scriptsize \textsc{Standard}} & {\scriptsize \textsc{User-exclusive}} & {\scriptsize \textsc{User-adaptive}} & {\scriptsize \textsc{Attribute discovery}} & {\scriptsize \textsc{Image clusters}}\\
 & & & & \citep{Kovashka13b} & \citep{Rastegari12} & \\
\hline
Pointy & \textbf{76.3} (0.3) & 74.0 (0.4) & 67.8 (0.2) & 74.8 (0.3) & 74.5 (0.4) & 74.3 (0.4)\\
Open & \textbf{74.6} (0.4) & 66.5 (0.5) & 65.8 (0.2) & 71.6 (0.3) & 68.5 (0.4) & 68.3 (0.4)\\
Ornate & \textbf{62.8} (0.7) & 56.4 (1.1) & 59.6 (0.5) & 61.1 (0.6) & 58.3 (0.8) & 58.6 (0.7)\\
Comfortable & \textbf{77.3} (0.6) & 75.0 (0.7) & 68.7 (0.5) & 75.5 (0.6) & 76.0 (0.7) & 75.4 (0.6)\\
Formal & \textbf{78.8} (0.5) & 76.2 (0.7) & 69.6 (0.4) & 77.1 (0.4) & 77.4 (0.6) & 77.0 (0.6)\\
Brown & \textbf{70.9} (1.0) & 69.5 (1.2) & 61.9 (0.5) & 68.5 (0.9) & 69.3 (1.2) & 69.8 (1.2)\\
Fashionable & \textbf{62.2} (0.9) & 58.5 (1.4) & 60.5 (1.3) & 62.0 (1.4) & 61.2 (1.4) & 61.5 (1.1)\\
\hline
Cluttered & \textbf{64.5} (0.3) & 60.5 (0.5) & 58.8 (0.2) & 63.1 (0.4) & 60.4 (0.7) & 60.8 (0.7)\\
Soothing & \textbf{62.5} (0.4) & 61.0 (0.5) & 55.2 (0.2) & 61.5 (0.4) & 61.1 (0.4) & 61.0 (0.5)\\
Open area & \textbf{64.6} (0.6) & 62.9 (1.0) & 57.9 (0.4) & 63.5 (0.5) & 63.5 (0.8) & 62.8 (0.9)\\
Modern & \textbf{57.3} (0.8) & 51.2 (0.9) & 56.2 (0.7) & 56.2 (1.1) & 52.5 (0.9) & 52.0 (1.1) \\
Rustic & \textbf{67.4} (0.6) & 66.7 (0.5) & 63.4 (0.5) & 67.0 (0.5) & 67.2 (0.5) & 67.2 (0.5)\\
\hline
\end{tabular}
\caption{Accuracy of predicting perceived attributes, with standard error in parentheses. Our shades provide robust models that capture personalized notions of the attributes, yet do not overfit to possible noise in a user's labels.}
\label{tab:shades_accuracy}
\end{table*}

\section{Experimental Validation}
\label{sec:results_shades}

We first demonstrate shades' key utility for improving attribute prediction (Section \ref{sec:schools-result}) and attribute-based image search (Section \ref{sec:shades_search}). 
We then quantitatively analyze the purity of the discovered shades (Section \ref{sec:coherency}).  
We offer comparisons to existing techniques, including both standard consensus attributes as well as state-of-the-art methods for attribute discovery \citep{Rastegari12} and personalized attributes \citep{Kovashka13b}.
We analyze shades qualitatively (Section \ref{sec:visualize}) to visualize what is discovered.
Finally, we show how to transfer shades between attributes and users in order to predict how a user will interpret an attribute for which he has provided no labels (Section \ref{sec:shades_tensor}).

\subsection{Implementation Details}
\label{sec:shades_exp_design}

We use image descriptors provided with the SUN and Shoes datasets for all methods: concatenated GIST and color histograms for Shoes, and GIST, color, HOG, and self-similarity histograms for SUN.  
See~\citep{Kovashka12,Patterson12} for details.
The datasets can be accessed at \url{http://vision.cs.utexas.edu/whittlesearch/} and \url{http://cs.brown.edu/}$\sim$\url{gen/sunattributes.html}, respectively.
We use the Bayesian Probabilistic Matrix Factorization (BPMF) implementation of~\cite{Xiong10}.  We fix $D=50$\footnote{See Figure \ref{fig:evalD} for an experiment on the sensitivity of our method to the choice of $D$.}, then use the default parameter settings.  For $N=1000$ and $M=195$, MCMC with 500 samples takes about 21 minutes.  We cross-validate all classifier parameters.  We set $K$ automatically per attribute based on the optimal silhouette coefficient within $K=\{2,\dots,15\}$.  Typically values of $K\approx7$ are chosen by the algorithm.  We evaluate all 12 attributes listed in Section \ref{sec:col}.

As noted in Section \ref{sec:col}, during data collection annotators must explain their attribute labels.  Specifically, we ask, ``Please explain your response.   What part or aspect of the image do you associate with the attribute [attribute name]? What part or aspect of the image led you to say that the attribute [attribute name] is present or not present?"  Figure~\ref{fig:descr} shows a sample of annotators' responses.  We draw on their explanations below to aid our quantitative evaluation, but they are never seen by our method.


\subsection{Accuracy of Perceived Shade Predictions}
\label{sec:schools-result}

We first demonstrate how well shades capture perceived attributes.  We apply the shades as described in Section \ref{sec:models} to predict user-specific labels.  We compare to five methods:

\begin{enumerate} 
\item \textsc{Standard}, which is the standard consensus approach used in \citep{Ferrari07,Kumar11,Lampert09,Farhadi09,Vaquero09,Branson10,WangY10,Patterson12}; 
\vspace{0.1in}
\item \textsc{User-exclusive}, which trains one attribute classifier per user using only his labeled images; 
\vspace{0.1in}
\item \textsc{User-adaptive}, a transfer method \citep{Kovashka13b} that adapts the majority vote model with the same user-specific labeled data as \textsc{User-exclusive}; 
\vspace{0.1in}
\item \textsc{Attribute discovery}, an alternative shade formation method that clusters images in the space of non-semantic attributes.  These attributes are splits in the feature space that are discriminative for object categories, and we find them with the state-of-the-art method of \cite{Rastegari12}\footnote{We use the code kindly provided by the authors; we train it with the 10 Shoes and 611 SUN categories in the training images used by our method. We also tried using the method of \cite{Rastegari12} with the semantic attributes as ``categories'', but it performed significantly worse.}; 
\vspace{0.1in}
\item and \textsc{Image clusters}, an additional alternative shade formation method inspired by prior work for discovering word ``senses'' (e.g., \cite{Loeff06}) that clusters the image descriptors for all images labeled positive by at least one annotator.
\end{enumerate}

For the last two baselines, in order to map an image cluster to ground truth descriptions, we look at the bag of images each annotator labeled as positive, find the image cluster to which the largest portion of the bag belongs, and assign it to be this user's shade ID.

All methods use linear SVMs for consistency with \cite{Kovashka13b}.  
Our method selects $K$ automatically per attribute, yielding values between 5 and 10.  We run 30 trials, sampling 20\% of the available labels to obtain on average 10 labels per user (representing what a user might reasonably contribute to train the system).

Table~\ref{tab:shades_accuracy} shows the results.  Our shade discovery method outperforms all other methods.  It is more reliable than \textsc{Standard}, which is the status quo attribute learning approach.
For ``open'', we achieve an 8 point gain over \textsc{Standard} and \textsc{User-exclusive}, which indicates both how different user perceptions of this attribute are, as well as how useful it is to rely on schools rather than individual users.  \textsc{Shades} also outperform the \textsc{User-adaptive} approach, while requiring the exact same labeling effort.  While that method learns personalized models, shades leverage \emph{common perceptions} and thereby avoid overfitting to a user's few labeled instances.  
For example, on ``brown'', \textsc{User-adaptive} actually decreases the accuracy of \textsc{Standard}, which shows that personalizing to individuals can be overkill as not every user has a unique perception. Rather, there are multiple shades of the attribute, and a user subscribes to some shade, hence \textsc{Shades}' superior performance.
Shades also outperform the two alternative shade formation baselines---\textsc{Attribute discovery} and \textsc{Image clusters}.
This shows that our approach for forming shades produces the highest-quality clusters which are most aligned with true user groupings based on the data provided, compared to other more ``obvious'' baselines.

\begin{figure}[t]
\includegraphics[width=0.48\linewidth]{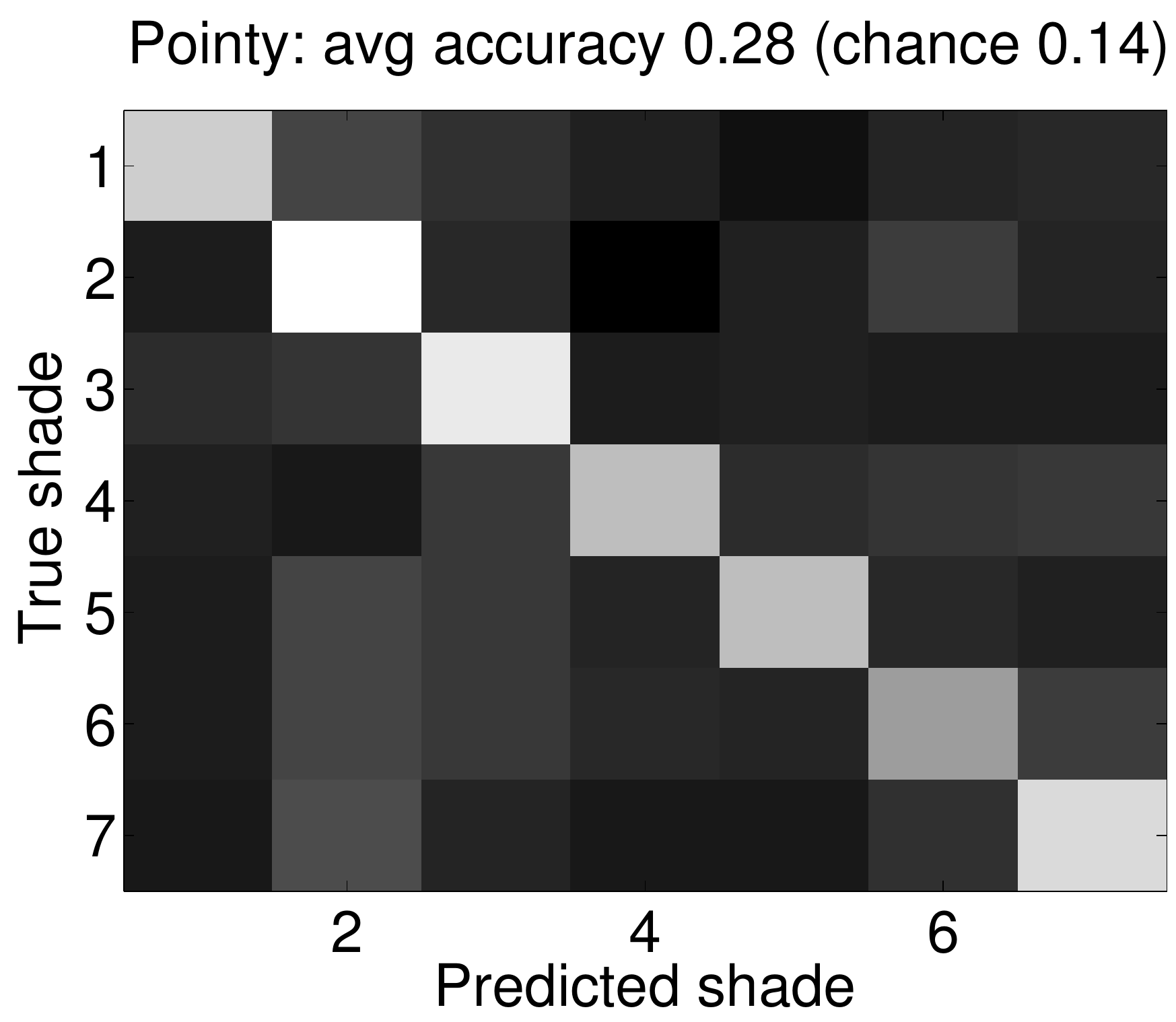} \includegraphics[width=0.49\linewidth]{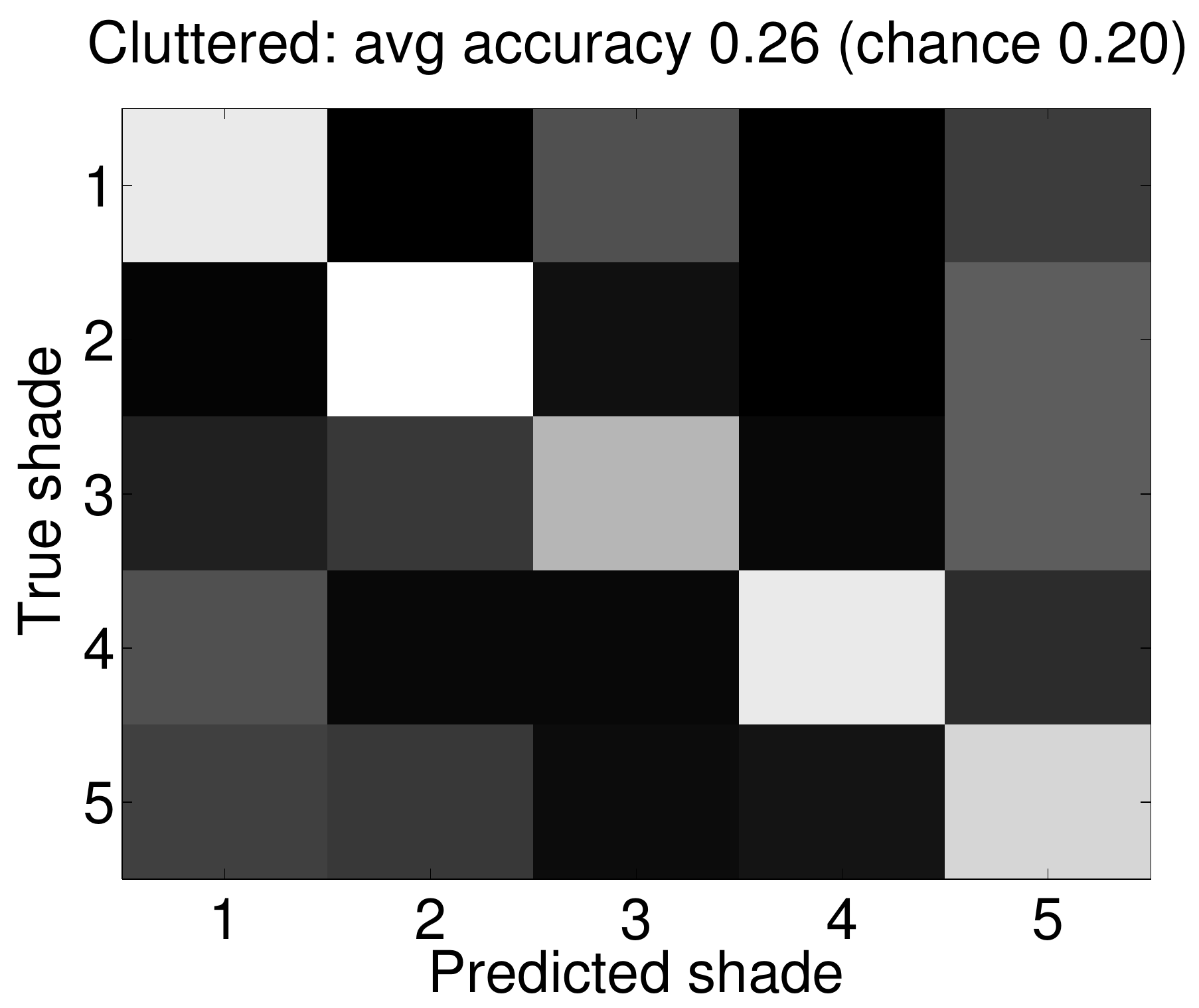}
\caption[Confusion matrices for multi-way shade classification]{Accuracy of perceived shade predictions:  Confusion matrices for multi-way shade classification, for the attributes ``pointy'' and ``cluttered''.}
\label{fig:confmat}
\end{figure}

While Table~\ref{tab:shades_accuracy} measures binary attribute classification, our method can also perform multi-way shade classification.  For this result, we cluster in the latent feature space of the images $I_j$, and again automatically select $K$.  Figure~\ref{fig:confmat} shows representative resulting confusion matrices for the attributes ``pointy'' and ``cluttered''.  Our average multi-way accuracy over all attributes is 0.28, much better than chance (0.15 on average).  This result indicates the discovered shades per attribute are indeed distinct and detectable.  

These results demonstrate the utility of shades. For all attributes, mapping a person's use of an attribute to a shade allows us to \emph{predict attribute presence more accurately}. This is achieved at no additional expense for each user.  As a result, applications demanding descriptive attributes (e.g., image search, zero-shot learning, etc.) can benefit from the more accurate representation.

\begin{figure}[t]
\includegraphics[width=1\linewidth]{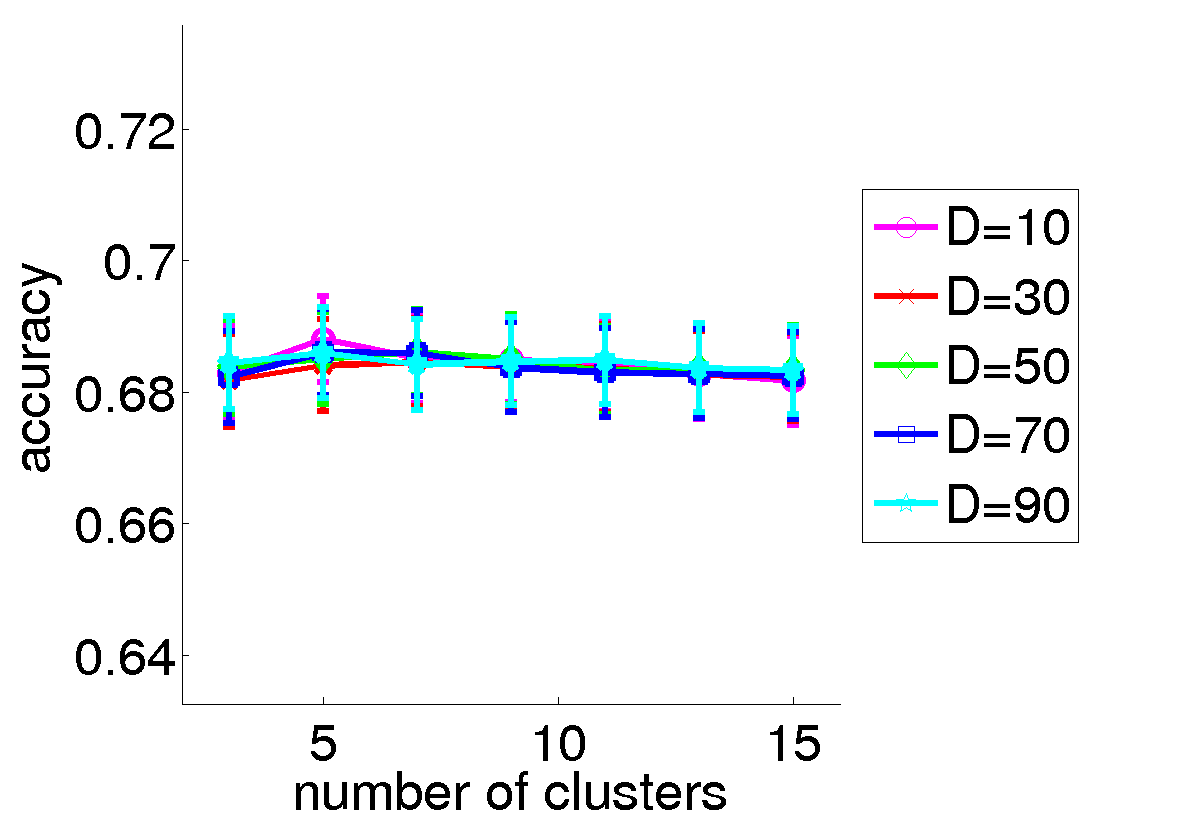}
\caption{Variance of shades' performance as a function of the number of latent factors $D$.}
\label{fig:evalD}
\end{figure}

Finally, we study the impact of the number of latent factors $D$ on the accuracy of attribute prediction with shades. 
In general, we can expect higher values of $D$ to enable better accuracy, whereas lower values of $D$ to allow faster computation. 
We run BPMF with 
$D = (10,100)$ in increments of 20.  
In Figure \ref{fig:evalD}, we plot attribute accuracy as a function of $D$, with varying values for $K$ (as the choice of $K$ might depend on the choice of $D$).  This figure shows an average over all attributes and 10 runs per attribute.  For 10 of the 12 attributes, the difference between accuracy scores is no more than 1\% depending on the choice of $D$, hence the small variance in the averaged plot.  Therefore, we conclude that our approach is not very sensitive to the choice of $D$.

\subsection{Personalized Image Search with Shades}
\label{sec:shades_search}

\begin{table*}[t]
\centering
\begin{tabular}{|c|c|c|c|c|c|}
\hline
$q$ & \textsc{Shades} & \textsc{Standard} & \textsc{User-exclusive} & \textsc{User-adaptive} & \textsc{Chance}\\
 & & & & \citep{Kovashka13b} & \\
\hline
2 & \textbf{53.3} (0.1) & 50.1 (0.1) & 43.3 (0.1) & 50.9 (0.1) & 25\\
\hline
3 & \textbf{39.8} (0.1) & 36.3 (0.1) & 29.4 (0.1) & 37.4 (0.1) & 12.5\\
\hline
4 & \textbf{29.7} (0.2) & 26.5 (0.2) & 20.1 (0.2) & 27.9 (0.2) & 6.2\\   
\hline
5 & \textbf{21.8} (0.5) & 18.8 (0.5) & 14.0 (0.4) & 20.7 (0.5) & 3.1\\   
\hline
6 & \textbf{17.1} (1.8) & 12.9 (1.6) & 11.7 (1.6) & 16.4 (1.8) & 1.5\\   
\hline
\end{tabular}
\caption{Multi-attribute query image search accuracy using shades, with standard error in parentheses.  $q$ is the number of attributes in the query.}
\label{tab:shades_search}
\end{table*}

Next we examine how the accurate perceived attribute models offered by shades can positively impact an image search application.

First, we collect additional data for the Shoes attributes in Table \ref{tab:defs}, such that the same images are labeled for all attributes, and all users label all attributes.\footnote{We omit the attribute ``brown'' since it only appears in a small set of images.} 
This is necessary since in the data collection described in Section \ref{sec:col}, many users only labeled a single attribute, so we have very few cases of multiple attributes labeled by the same user for the same image.
We ask each of 200 users to label 40 images for each attribute, out of a total set of 200 images that receive labels from any user. We use 50 images total for training, 75 for testing, and 75 for cross-validation. We repeat the shade formation and shade-based attribute prediction procedure as in Section \ref{sec:models}, using the training data from each user. 

We then pose multi-attribute queries with the test images.  For each test image and user, we generate all $q$-tuples of the attributes with labels from the user. Each of these tuples forms a multi-attribute query composed of $q$ attributes that a user might issue during search, e.g., ``I want to buy \emph{ornate} and \emph{formal} shoes.'' We use the user's labels as the ground truth for these queries, and examine the presence/absence predictions of the \textsc{Standard}, \textsc{User-exclusive}, \textsc{User-adaptive}, and \textsc{Shades} approaches on each $q$-attribute query.  
To quantify retrieval accuracy, we measure the fraction of these query images where the user's ground truth labels and a model's predictions agree on all $q$ attributes per query.

\begin{figure*}[t]
\includegraphics[width=1\linewidth]{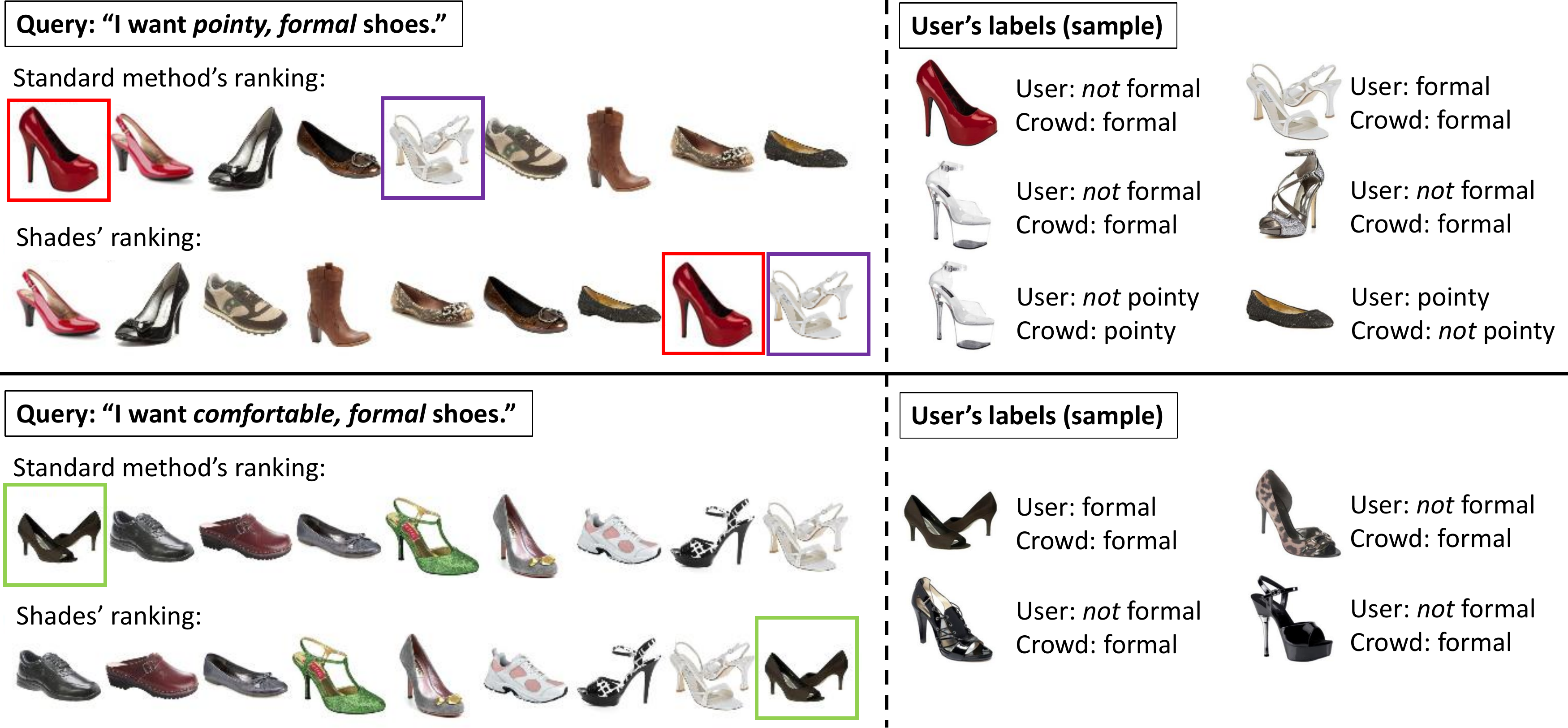}
\caption{Qualitative result of image search using shade models as opposed to standard attribute learning models.  Our shades retrieve results which more accurately capture the user's notion of the attributes, without overfitting to individual labels.  See the text for more details.}
\label{fig:qual_search}
\end{figure*}

Table \ref{tab:shades_search} shows the results, for $q=\{2, \dots, 6\}$. 
Our shades approach produces higher match rates, hence more accurate image search results, than any of the baselines, consistent with our result in Section \ref{sec:schools-result}. 
For $q=2$, our method achieves a 6\% relative gain over \textsc{Standard}, and 5\% gain over \textsc{User-adaptive}.
This demonstrates that in order for attribute-based searches to be successful, the retrieval system needs to interpret the user's attribute queries correctly; shades allow the learning of robust models which are personalized yet do not overfit to noise in a user's labels.

Note that chance performance corresponds to the probability of randomly matching all $q$ attribute ground truth labels.  
All methods show a decrease in accuracy as more query words are used, since it becomes more difficult for a method to correctly predict the presence of \emph{all} increasingly many attributes.

Figure \ref{fig:qual_search} shows a qualitative search result.  We rank the subset of database images for which we have user labels based on how many of the requested attributes they are predicted to have, for both the \textsc{Standard} approach and our \textsc{Shades} approach.  We also show a subset of the user's labels as well as the majority-voted labels for the same image, which helps explain the result.  For the first query, notice how our method ranks the red stiletto shoe (outlined in red) compared to the baseline.  Our method observes the user's idea that shoes with very high heels are neither ``formal'' nor ``pointy'' (first column of user labels).  Further, even though the user agreed with the crowd on the ``formalness'' of the sandal shoe outlined in purple, he rated other open shoes as ``\emph{not} formal'', so our shades model correctly learned that sandals should be ranked low given a query for ``formalness''.  For the second query example, notice that even though the user agreed with the crowd regarding the ``formalness'' of the shoe outlined in green, he labeled other similar-looking shoes as ``\emph{not} formal''.  Our shades model captures this trend, rather than overfitting to an individual user label, and ranks the green-outlined image low.


\subsection{Quantifying the Accuracy of Shade Formation}
\label{sec:coherency}

\begin{figure}[t]
\centering
\includegraphics[width=1\linewidth]{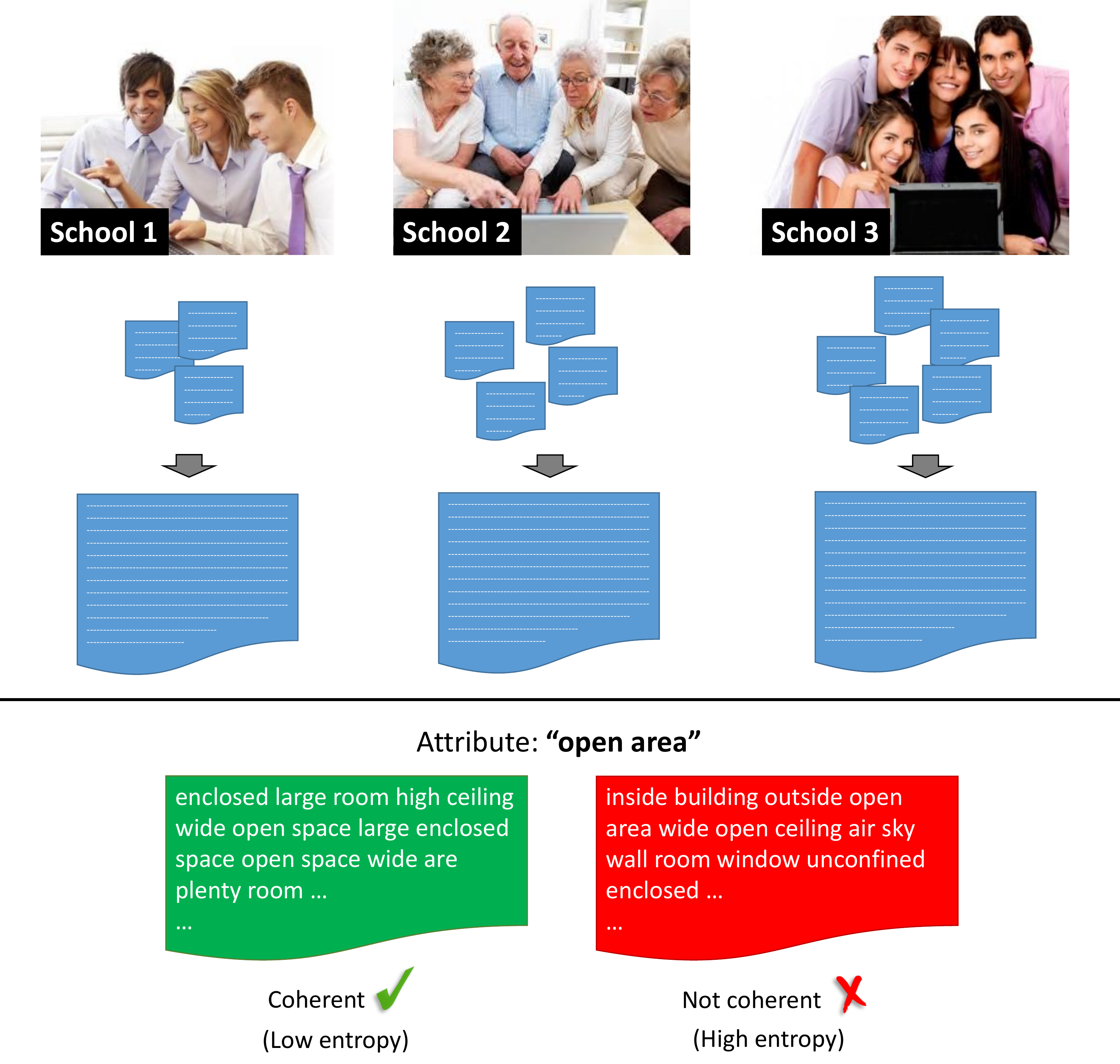}
\caption{Illustration of our cluster coherency evaluation. Top: We pool together the label explanations from each user in a school, and then examine the distribution over topics for each per-school document. Bottom: A coherent document is one that focuses on just a few topics (e.g., ``open areas'' which are inside, in this case) as opposed to many topics (e.g., both inside and outside ``open areas'').}
\label{fig:coh_expl}
\end{figure}

To further quantify how accurately our shades capture perceived interpretations, we next score how \emph{coherent} the textual explanations (cf. Figure \ref{fig:descr}) are among annotators in the same shade.  
In particular, we quantify how coherent the label explanations are when we pool the text from all users within a given shade.  See Figure \ref{fig:coh_expl}.  
Whereas random clusters would group diverse ground truth explanations together, good shades should align with coherent explanations.  We stress that these explanations are never seen by our algorithm; they are for evaluation purposes only.

To measure coherency, we use a text analysis metric for \emph{topic entropy}~\citep{Hall08}. 
We first perform probabilistic Latent Semantic Analysis (pLSA)~\citep{Hofmann99} on the Porter-stemmed textual descriptions.  We treat each description for which $L_{ij}=1$ as a document and discover $T=200$ topics with pLSA.  Then we map each explanation to its distribution of topics (a vector of $T$ weights).  This representation accounts for word meaning, not just word occurrences  (e.g., ``image'' and ``picture'' will be treated as synonyms by pLSA).  
Let $\mathbf{W}^k$ denote the matrix whose columns are the $T \times 1$ topic representation vectors for each of the $V$ positive explanations corresponding to users in shade $\mathcal{S}_k$. We define the representation of topics in this shade as $q^k = \frac{1}{V} \sum_{v=1}^V \mathbf{W}^k_{:,v}$, where the index $:,v$ denotes a column of the matrix. Then we compute the overall topic entropy for this shade as $- \sum_{t=1}^T q^k_t \log q^k_t$.  
Low entropy is better, as it indicates the shade corresponds to a more coherent set of descriptions focused around a few topics. 

We compare \textsc{Shades} to two methods defined above in Section \ref{sec:schools-result}: 
\begin{enumerate}
\item \textsc{Attribute discovery}: the state-of-the-art non-semantic attribute discovery method of \cite{Rastegari12}; and
\item \textsc{Image clusters}: an image clustering approach inspired by \cite{Loeff06}.  
\end{enumerate}

These baselines represent how one might reasonably attempt to perform shade formation with existing techniques.

Note that all methods use $K$-means and remove clusters with fewer than 10 members, which tend to be too sparse to form a meaningful shade.

\begin{figure}[t]
\centering
\includegraphics[width=1\linewidth]{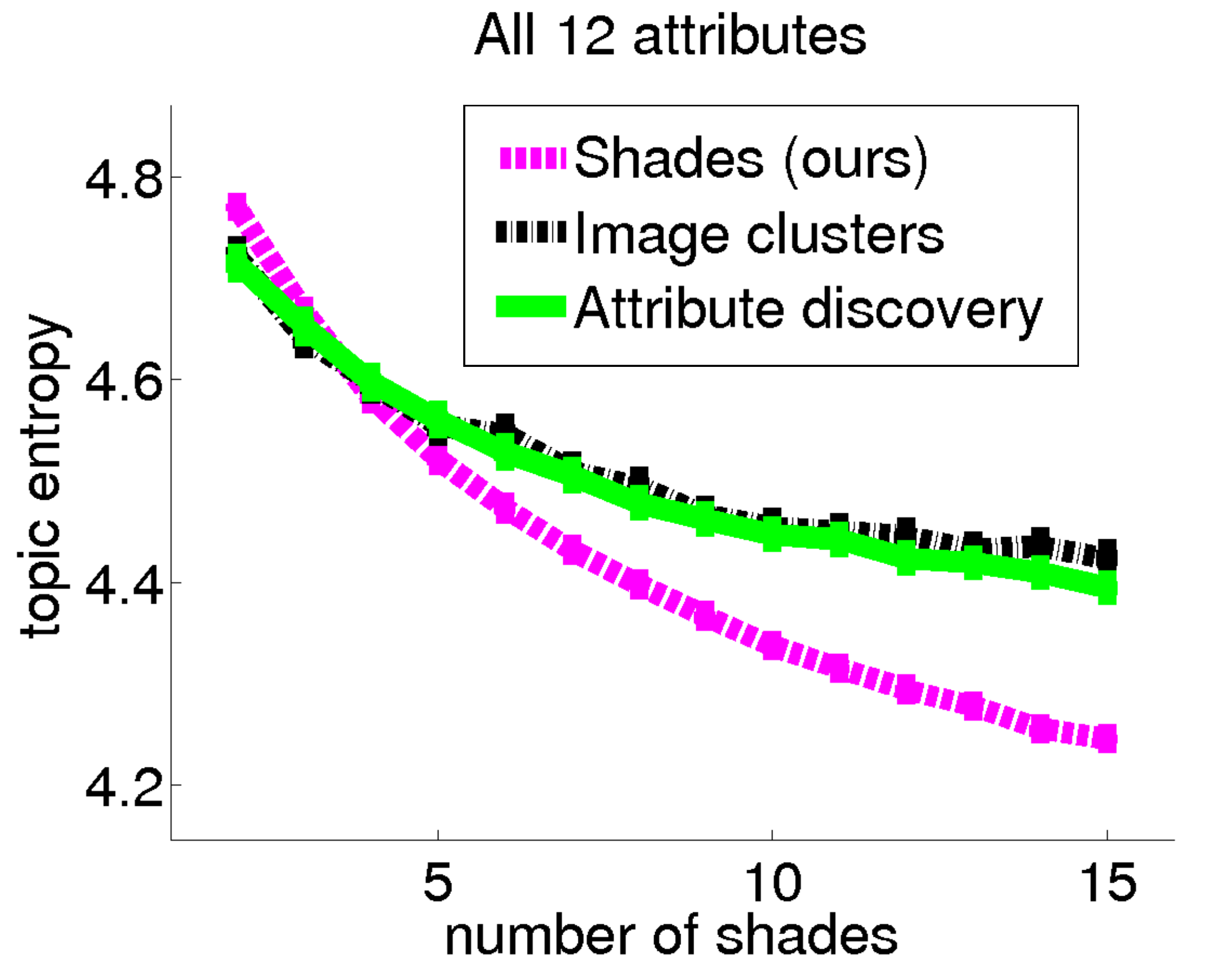}
\caption{Quality of discovered attribute shades (low entropy indicates a more coherent shade/cluster).}
\label{fig:nlp_eval}
\end{figure}

Figure~\ref{fig:nlp_eval} shows the results.  We plot topic entropy (and standard error) as a function of the number of shades $K$, over all attributes and 30 runs.  
Our shades are much more coherent overall.  Clearly, image clustering falls short.  The non-semantic attribute discovery method of \cite{Rastegari12}, while stronger than clustering, does not capture the shades of meaning since it lacks human input on the attribute interpretation.  When $K=2$, the baselines have lower entropy than our shades, showing that very coarse groups are sufficiently found with image clustering; however, these clusters are too coarse according to the silhouette coefficient model selection, which selects $K=5$ to $K=10$ shades as the optimal setting. This shows the shades we have discovered are meaningful and accurately capture the varied attribute meanings that users employ.

We now give some more information to help gauge the significance of these results.  Our method achieves entropy which is about 0.2 lower than the entropy of the baseline methods.  In Table \ref{tab:ent}, we show some pairs of individual descriptions which have about 0.2 difference in their topic distribution entropies.\footnote{Note that Figure \ref{fig:nlp_eval} captures entropies of distributions over a number of descriptions, which are naturally higher than the topic entropy of a single description.}  Again, lower entropy denotes a more focused explanation. In Table \ref{tab:ent}, the first explanation for ``open'' includes many unrelated details, while the second predominantly discusses the foot being seen.  Similarly, a high-quality user cluster will correspond to explanations that focus on a single or a few topics.  The second explanation for ``ornate'' focuses on color, hence achieves lower entropy.  The second explanation for ``open area'' focuses on the words ``room'' and ``space''. Just like the second explanation in each pair, the clusters that our method obtains are more focused.

\begin{table}[t]
\centering
\begin{tabular}{|c|c|p{4.6cm}|}
\hline
Attribute & Entropies & Explanations\\
\hline
Open & 2.85 & ``This shoe is open across the top of the foot, with a space between the ankle strap and the toe. It also has gaps along the sides of the toe.''\\
     & 2.62 & ``Open represents that amount of foot that can be \textbf{seen} when the show is worn. The opening on this shoe allows for a portion of the upper foot to be \textbf{seen}.''\\
\hline
Ornate & 2.45 & ``I consider the shoe in Image 45 to be ornate (made in an intricate shape or decorated with complex patterns) because it is oddly shaped, with a pattern and added strapping and it has a zipper pull that stands out.''\\
       & 2.27 & ``I associate the pattern of the shoe with the attribute ornate. It is the way that the plaid is mixed in, its \textbf{color}, and the mixing of the \textbf{color} in the shoe laces as well that led me to say that the attribute ornate is present.''\\
\hline
Open area & 2.41 & ``You can see the sky and even though the photo is of a building there is plenty of open space surrounding it as well as the photography being taken outside.''\\
          & 2.23 & ``Inside the net there is plenty of \textbf{space}, and \textbf{room} between the nets. There's not too much \textbf{room}, but enough to be considered an open area. It's also outside so out of the nets is plenty of \textbf{room}.''\\
\hline
\end{tabular}
\caption{Pairs of annotation explanations with corresponding topic entropy.  Bold is our emphasis.  Notice how lower entropy corresponds to more focused description (second example in each attribute).  Similarly, our shades method produces more focused clusters.  See the text for an explanation.}
\label{tab:ent}
\end{table}


\subsection{Visualizing Attribute Shades of Meaning}
\label{sec:visualize}

\begin{figure*}[htp]
\centering
\includegraphics[width=1\linewidth]{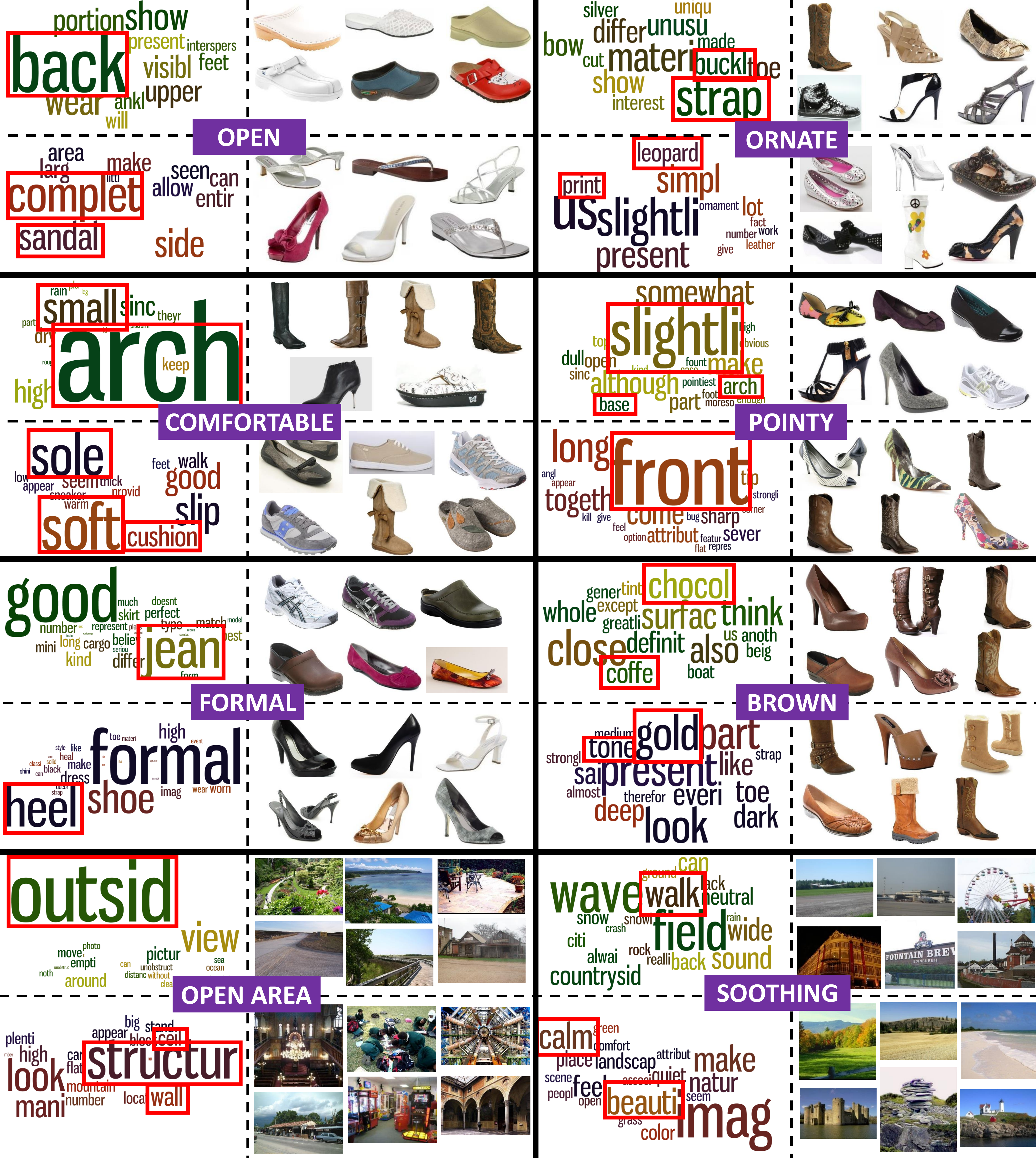}
\caption{Top words and images for two shades per attribute (top and bottom for each attribute).  Best viewed on PDF or in color.  Notice the subtle differences in the annotator notions of the attributes exemplified by both the images considered positive for each shade, as well as the most frequent words in the corresponding textual explanations.  See the text for a more detailed description.}
\label{fig:shades}
\end{figure*}

Next, we provide qualitative results.  Figure \ref{fig:shades} visualizes two shades each, for eight of the attributes.  The images are those most frequently labeled as positive by annotators in a shade $\mathcal{S}_k$.  The (stemmed) words are those that appear most frequently in the annotator explanations (cf.~ Figure~\ref{fig:descr}) for that shade, after we remove words that overlap between the two.  Font size reflects relative frequency.  To aid readability, we also outline words that stand out as good representatives of the shade.
Recall that the text annotations are \emph{not} used by our approach during shade discovery.

We see the shades capture nuanced visual sub-definitions of the attribute words.  
For example, for the attribute ``brown", one shade covers chocolate-colored shoes (top shade), while another is lighter and more gold (bottom shade).  
For ``ornate", one shade focuses on straps/buckles (top), while another focuses on texture/print/patterns (bottom).  
For ``comfortable'', one shade emphasizes a low arch (top), while the other requires soft materials (bottom).
For ``pointy", one focuses on the front of the shoe (bottom), while another focuses on heels/bases that are ``slightly" pointy.  
For ``open'', one shade includes open-heeled shoes, while another includes sandals which are open at the front \emph{and} back.
In SUN, the ``open areas" attribute can be either outside (top) or inside (bottom).  
For ``soothing", one shade emphasizes scenes conducive to relaxing activities, while another focuses on aesthetics of the scene.


As discussed above, an important feature of our method is its ability to perform discovery independent of a particular image descriptor.  To illustrate this, we next use the shades' visual classifiers to examine their most informative \emph{localized} features.  We use $L_1$ regularization when training one-vs.-rest logistic regression classifiers for each shade, in order to isolate a sparse set of features most discriminative for that shade.  For each 70 $\times$ 70 grid cell of the image, we sum the magnitude of the classifier weights for its features.  Then we multiply those weights with the pixel intensities in order to visualize the relative impact of each portion of the image.

\begin{figure}[t]
\includegraphics[width=1\linewidth]{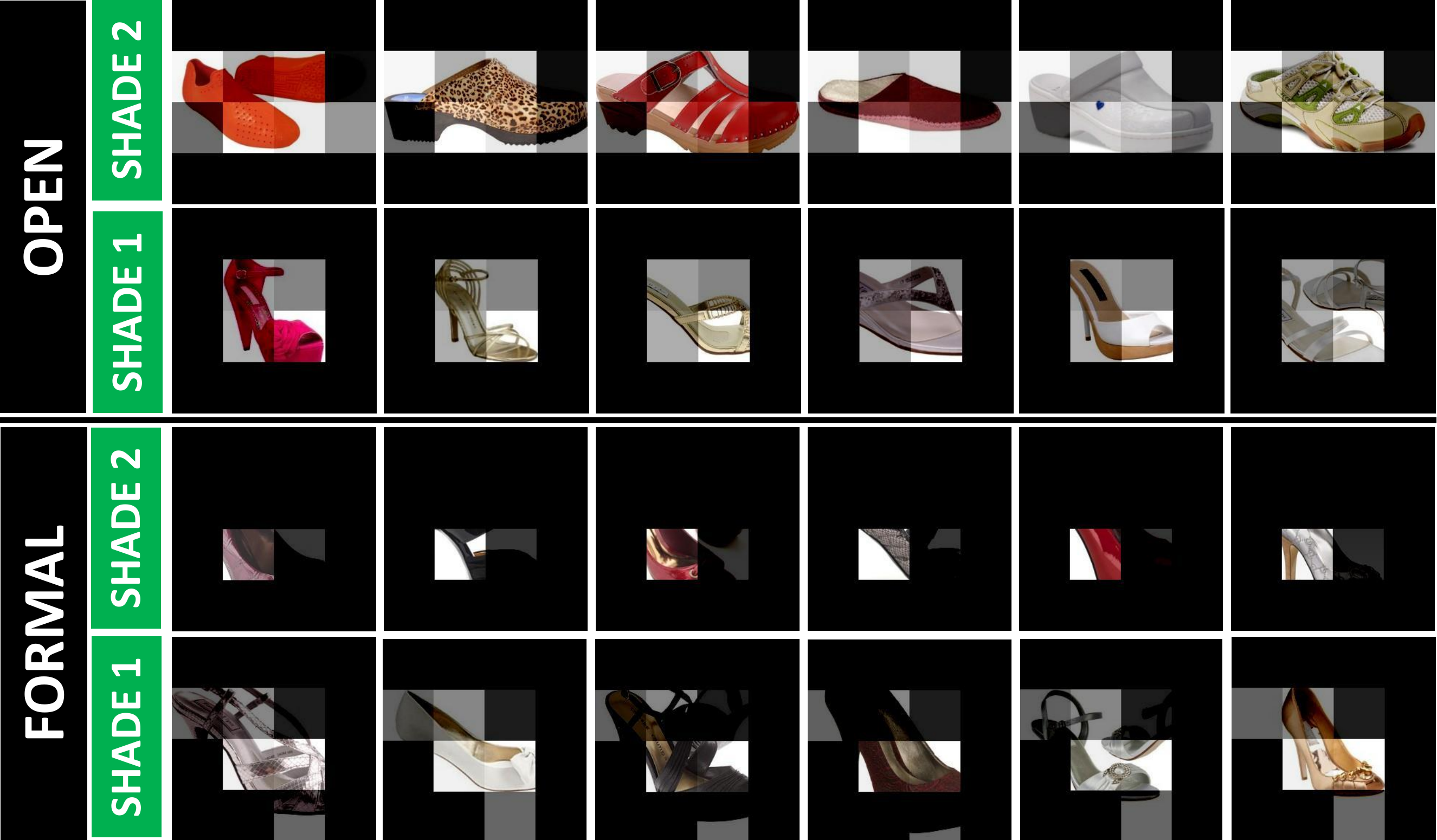}
\caption[Image regions that are important for learning the attribute shades]{Image regions highlighted according to the importance of the localized features for learning the shades.  Our method finds those localized visual properties that determine whether a shaded attribute is present or not.}
\label{fig:shoe_regions}
\end{figure}

Figure~\ref{fig:shoe_regions} shows example results.   Brighter cells indicate regions more discriminative for that shade.  For ``open", we see one shade emphasizes openness at the back, and another openness at the toe.  For ``formal'', the top shade emphasizes the arch of the shoe, while the bottom one emphasizes the toes.  Such examples illustrate how our method isolates visual properties that support a shade, yet would not be tightly grouped if simply clustering global descriptors.

Of course, learning discriminative spatially localized features is nothing new; our point is that shades are what enable the training image groups that make this discriminative selection feasible.  Furthermore, recent work using crowds to isolate informative spatial regions \citep{Donahue11,Deng13} has a different purpose (fine-grained image classification) and takes an entirely different approach (explicitly asking labelers to outline the regions needed to make their label decisions).


\subsection{Exploiting Attribute Correlations for Cross-Attribute Transfer}
\label{sec:shades_tensor}

So far, we have discovered the shades of each attribute disjointly from other attributes. However, the attributes that we use are not completely independent. For example, there is notable correlation between the attributes ``fashionable'' and ``formal''.  
We propose to exploit these correlations to predict how a user will perceive an attribute for which he has not supplied any labeled examples, by transferring labels for this attribute from other users, and from other attributes labeled by the same user.

As mentioned in Section \ref{sec:schools}, matrix factorization can also be used to ``fill in'' missing values in the (user, image) label space. The value of an entry $L_{ij}$ can be computed as an inner product of the user $A_i$'s and image $I_j$'s latent factor vectors. 

However, this label imputation can also exploit \emph{multiple} (user, image) label matrices together, if we stack these matrices in a tensor.
In this case, the label matrix $\mathbf{L}$ becomes an $M \times N \times Z$ label \emph{tensor}, where $Z$ denotes the number of attributes being considered at once.
We can decompose $\mathbf{L}$ as:
\begin{equation}
\mathbf{L} = \sum_{d=1}^D \mathbf{A}_{d, :} \circ \mathbf{I}_{d, :} \circ \mathbf{T}_{d, :},
\end{equation} 
where, the index $d, :$ refers to the rows of the matrices and $\circ$ refers to outer vector product. 
$\mathbf{T}$ is the $D \times Z$ matrix of latent factors for each of the $Z$ attributes.
We use the Bayesian tensor factorization of \cite{Xiong10} for this formulation, which essentially extends the probabilistic matrix factorization approach of Salakhutdinov and Mnih discussed above to handle tensor data.

An entry $L_{ijz}$ denotes how user $i$ labeled image $j$ for attribute $z$. Equation \ref{eqn:shades} then becomes

\begin{equation}
p(\mathbf{L} | \mathbf{A}, \mathbf{I}, \mathbf{T}, \sigma^2) = \prod_{i=1}^M \prod_{j=1}^N \prod_{z=1}^Z \big[\mathcal{N}(L_{ijz} | \langle A_i, I_j, T_z \rangle, \sigma^2)\big]^{\ell_{ijz}},
\end{equation}
where $A_i$ and $I_j$ denote columns of $\mathbf{A}$ and $\mathbf{I}$ as before, $T_z$ denotes a column of $\mathbf{T}$, 
and we model the prior over the latent factors in $\mathbf{T}$ as a spherical Gaussian, similar to $\mathbf{A}$ and $\mathbf{I}$.
See the Bayesian Probabilistic Tensor Factorization (BPTF) approach of \citep{Xiong10} for more details.

\begin{table}[t]
\centering
\begin{tabular}{|c|c|c|}
\hline
Dataset & Ours & Chance\\
\hline
Shoes & 0.831 (0.001) & 0.50\\
SUN & 0.770 (0.001) & 0.50\\
\hline
\end{tabular}
\caption{Accuracy of imputing missing labels using other attributes, with standard error in parentheses. Utilizing attribute correlations allows us to accurately predict how a user will perceive a novel attribute, without having received any annotations for this attribute from this user.}
\label{tab:tensor}
\end{table}

Using this tensor label imputation approach, we can complete a transfer learning task of predicting how a user who has never labeled an attribute $z$ will perceive this attribute, by relying on this user having labeled other attributes, and other users having labeled attribute $z$. 

Table \ref{tab:tensor} shows the results.
For Shoes, we use the new data collected in Section \ref{sec:shades_search} as it ensures all users have labeled all attributes, while for SUN we lack such data and use the data collected in Section \ref{sec:col}.
We achieve a much higher accuracy than chance performance at 50\%, thus showing that one can successfully transfer knowledge about one attribute to another.

\section{Conclusion}

Our work addresses the gap between how people \emph{describe} attributes and how they \emph{perceive} them visually.  
We show how to discover people's shared biases in perception, then exploit them with visual classifiers that can generalize to new images.  
The proposed approach to discover attribute shades brings together language, crowdsourcing, human perception, and visual representations in a new way.

The learned shades successfully tailor attribute predictions to cater to a user's ``school of thought'', boosting the accuracy of detecting perceived attributes.  
In systematic experiments, we quantify the impact of shades, both compared to standard paradigms and multiple state-of-the-art methods.  
We demonstrate that for image search applications, it is crucial to build robust personalized models that account for a user's biases.
The visualized shades show great promise to separate the (sub-)attributes involved in a person's use of an attribute vocabulary during image search or organization of image content.

It is plausible that shades originate in part due to cultural differences that might be captured well by demographic information, like a person's location, age, etc.  We conducted a preliminary study to determine whether shades correlate with demographics.  We asked some annotators from the United States to name their city of residence, and after performing clustering in latent factor space, 
we mined for correlations between the clusters found and the annotators' geographic locations.
However, clusters in the latent factor space did not produce obvious clusters in geographic space.  This suggests that shades are more subtle than what is captured within demographic parameters alone.  This problem merits further exploration, including by extending the range of the study to countries other than the US.

In future work, we will investigate ways to predict a person's preferred shade based on a minimal set of label requests.
We would also like to further explore the semantic relationships between the attributes, to determine how transfer across attributes might help learn shade models more efficiently.
Additionally, we would like to study approaches for automatically determining the degree of ambiguity in an attribute term from the attribute's textual definition, possibly with the addition of a small number of image exemplars.
Finally, it would be intriguing to apply our approach for novel tasks, such as discovering the common types of errors annotators make (for purposes of illustration during training) and for examining ambiguity in descriptions of actions.


\begin{acknowledgements}
We thank the anonymous reviewers for their helpful feedback and suggestions.  This research is supported in part by ONR ATL N00014-11-1-0105.
\end{acknowledgements}

\bibliographystyle{spbasic}      
\bibliography{thesis_refs_alpha}   

\end{document}